\documentclass{article}
\usepackage{iclr2027_conference,times}

\usepackage{amsmath}
\usepackage{amssymb}
\usepackage{booktabs}
\usepackage{longtable}
\usepackage{float}
\usepackage{placeins}
\usepackage{needspace}
\usepackage{framed}
\usepackage{fvextra}
\usepackage{graphicx}
\usepackage{microtype}
\usepackage{multirow}
\usepackage{array}
\usepackage{xcolor}
\usepackage{tikz}
\usetikzlibrary{arrows.meta}
\usepackage{hyperref}
\usepackage{url}
\hypersetup{hidelinks}

\newcommand{\system}{\textsc{SkillScriptBench}}

\title{\raggedright SkillScriptBench: Benchmarking Self-Evolution of Executable Agent Skill Packages Beyond Markdown}

\author{\normalfont
Yuxuan Liu\textsuperscript{1},
Haoran Li\textsuperscript{1}\thanks{Corresponding author.},
Yuhao Zhang\textsuperscript{3},
Jiahe Guo\textsuperscript{2},\\
\normalfont
Hongyu Luo\textsuperscript{1},
Wenbin Hu\textsuperscript{1},
Huihao Jing\textsuperscript{1},
Kawai Chung\textsuperscript{1},
Junle Chen\textsuperscript{1},\\
\normalfont
Changxuan Fan\textsuperscript{1},
Qing Zong\textsuperscript{1},
Lingyun Xie\textsuperscript{1},
Yangqiu Song\textsuperscript{1}\\
\normalfont\small
\textsuperscript{1}The Hong Kong University of Science and Technology\\
\normalfont\small
\textsuperscript{2}Harbin Institute of Technology, Harbin, China\\
\normalfont\small
\textsuperscript{3}Harbin Institute of Technology, Shenzhen, China\\
\normalfont\small
\texttt{\{yliurk,hlibt\}@connect.ust.hk}, \texttt{yqsong@cse.ust.hk}}

\iclrfinalcopy
\begin{document}
\setcounter{footnote}{1}
\maketitle
\fancyhead{}

\begin{abstract}
Executable Agent Skills combine natural-language instructions and scripts into reusable packages for LLM agents, and revising them requires fixing errors without breaking correct behavior. Existing benchmarks do not systematically distinguish documentation repair, script repair, and preservation when evaluating skill self-evolution. We introduce SkillScriptBench, a 350-task benchmark designed to evaluate these capabilities separately. From a survey of over 35,000 GitHub-hosted Skill roots, we select 100 packages and construct 150 repair tasks. Each task pairs a package containing injected script faults with a maintenance request and executable checks of the required behavior. A complementary controlled track contains 200 tasks from 50 packages, each evaluated under the same maintenance request in four states: clean, documentation faults, script faults, and faults in both. Across four LLMs, methods that edit both documentation and scripts can repair script faults but do not consistently outperform Markdown-only revision on documentation repair or preservation. We therefore introduce AST-Guided Skill Revision, which uses abstract syntax trees and calling relationships to link maintenance requirements to relevant code locations. It restricts script edits to these locations and updates the documentation to match the revised scripts. Averaged across models, this revision stage yields absolute gains in repair success of 21.9\% for Raw Package and 27.7\% for CoEvoSkills on faulty packages. Absolute gains in the proportion of tasks solved in all three runs reach 20.8\% and 31.5\%, respectively, indicating more consistent repair success across repeated runs.
The benchmark and evaluation tools are available at
\url{https://github.com/KnowComp/SkillScriptBench}.
\end{abstract}

\section{Introduction}
\label{sec:introduction}

LLM agents increasingly perform multi-step tasks through sustained interaction with tools, files, and execution environments~\citep{yao2023react,yang2024sweagent}. Reliable execution depends on procedural knowledge for coordinating operations, recovering from failures, and validating outputs~\citep{wang2023voyager,shinn2023reflexion}. Agent Skills make this knowledge reusable by organizing instructions in a \texttt{SKILL.md} file, optionally accompanied by executable scripts and supporting resources~\citep{agentskills_spec_2026,li2026skillsbench}. These packages provide workflows that agents can reuse across tasks, while skill self-evolution uses experience and feedback to revise them as requirements change or failures emerge~\citep{zhong2026skilllearnbench,zhang2026coevoskills,liu2026skillrevise,liu2026rethinking}. For executable skills, this process involves both the documented workflow and its implementation, making consistency between documentation and scripts an important part of reliable evolution.

Existing research spans three settings: evaluating the downstream utility of fixed skills, generating reusable skills from task experience or source corpora, and refining existing skills through feedback. Fixed-skill benchmarks assess packages without revising them, while representative skill-generation and continual-learning benchmarks often use a \texttt{SKILL.md} document as the learned artifact~\citep{li2026skillsbench,han2026sweskillsbench,zhong2026skilllearnbench,zhou2026skillgenbench}. Among self-evolution methods, SkillOpt and SkillAudit primarily revise instruction documents, whereas CoEvoSkills and SkillEvolver support executable multi-file packages~\citep{yang2026skillopt,gao2026skillaudit,zhang2026coevoskills,zhang2026skillevolver}.

Despite this progress, the self-evolution of existing executable skill packages remains underexplored in current benchmarks, particularly the revision of scripts and their consistency with documentation. These components can fail independently: a documented command may pass an unsupported option to an otherwise correct script, while a script may mishandle an option that is correctly documented. Task success and broader skill-quality assessments do not, by themselves, distinguish documentation repair from script repair, and aggregate results can obscure regressions in packages that were already correct~\citep{li2026skillsbench,zhong2026skilllearnbench,yang2026skillopt,zhang2026coevoskills}. Addressing these gaps requires a benchmark that evaluates revisions for fault repair, documentation--script consistency, and preservation of correct behavior across diverse executable packages.

\begin{figure}[!t]
\centering
\setlength{\abovecaptionskip}{4pt}
\makebox[\linewidth][c]{\includegraphics[width=1.04\linewidth]{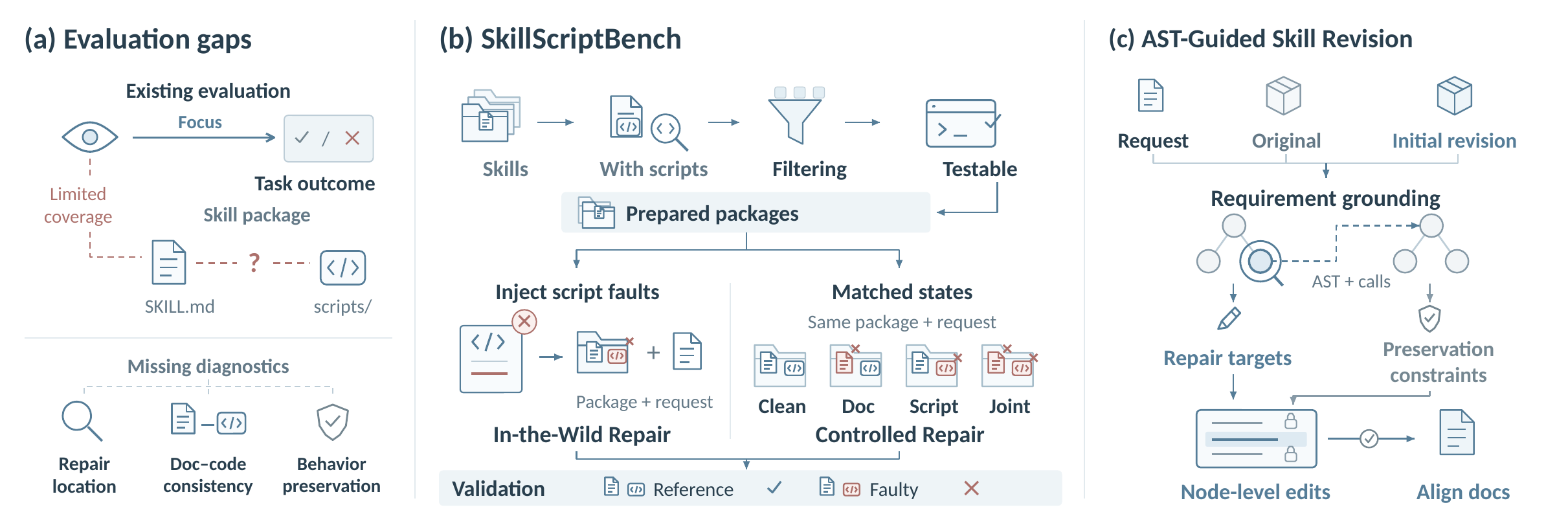}}
\caption{Overview of \system{} and AST-Guided Skill Revision: evaluation gaps,
benchmark construction, and coordinated documentation--script revision.}
\label{fig:benchmark-overview}
\vspace{-10pt}
\end{figure}

To address these gaps, we introduce \system{}, a 350-task benchmark that treats existing executable Agent Skill packages as the unit of self-evolution and evaluation. Each task pairs a package with a maintenance request. For the In-the-Wild Repair track, we scan over 35,000 GitHub directories containing \texttt{SKILL.md} and construct 150 repair tasks by injecting script faults into 100 curated packages. The Controlled Repair track contains 200 tasks from 50 packages, each provided in clean, documentation-fault, script-fault, and joint-fault states with its maintenance request held fixed. Together, the two tracks cover varied package structures, programming languages, and maintenance requirements while supporting controlled comparisons of repair and preservation. Figure~\ref{fig:benchmark-overview} summarizes the benchmark and revision framework.

Across four LLMs, methods that revise both documentation and scripts can repair script faults but do not consistently outperform Markdown-only revision on documentation repair or clean-package preservation. To better coordinate repair and preservation, we introduce AST-Guided Skill Revision as an additional stage for existing revision methods. It analyzes the original package and an initial revision, links maintenance requirements to abstract syntax tree (AST) nodes, and uses calling relationships to provide context for targeted edits. Script edits are restricted to selected nodes and guided by preservation constraints derived from requirements assessed as satisfied. The documentation is then aligned with the maintenance request and revised implementation. Averaged across the four models, this stage increases repair success on faulty packages from 54.8\% to 76.7\% for Raw Package and from 48.8\% to 76.5\% for our CoEvoSkills adaptation. The proportions of tasks solved in all three runs increase by 20.8 and 31.5 percentage points, respectively, and both augmented methods improve clean-package preservation relative to their corresponding baselines.

\textbf{Contributions.}
Our contributions are threefold.
\par\noindent\textbf{1. A benchmark for executable skill self-evolution.}
We introduce \system{}, a 350-task benchmark that treats existing Agent Skill
packages as the unit of evolution, covering documentation, scripts,
and their consistency through repository-sourced tasks and matched clean and
faulty states.
\par\noindent\textbf{2. Empirical findings on repair and preservation.}
Across four LLMs, we show that full-package revision supports script repair but
does not consistently improve documentation repair or clean-package preservation
over Markdown-only revision.
\par\noindent\textbf{3. A structure-guided self-evolution method.}
AST-Guided Skill Revision combines targeted script edits, preservation
constraints, and documentation alignment, improving repair and preservation
when applied to Raw Package and adapted CoEvoSkills outputs.

\section{SkillScriptBench}
\label{sec:benchmark}

\system{} evaluates how Agent Skills revise their Markdown instructions and
executable scripts to satisfy maintenance requests, using 350 tasks across
two complementary tracks. In-the-Wild
Repair covers diverse public packages and maintenance requirements. Controlled Repair uses matched clean and faulty package states to assess documentation
repair, script repair, joint repair, and preservation of correct behavior.
Table~\ref{tab:benchmark-comparison} compares the package-maintenance coverage of
\system{} and representative Agent Skill benchmarks~\citep{li2026skillsbench,
han2026sweskillsbench,zhong2026skilllearnbench,zhou2026skillgenbench}.

\begingroup
\setlength{\intextsep}{8pt}
\begin{table}[H]
\caption{Comparison of Agent Skill benchmarks.\newline $\checkmark$ / $\times$ indicate
whether each dimension is explicitly covered.}
\label{tab:benchmark-comparison}
\centering
\small
\setlength{\tabcolsep}{3pt}
\renewcommand{\arraystretch}{1.12}
\begin{tabular*}{\linewidth}{@{\extracolsep{\fill}}lcccc@{}}
\toprule
Benchmark &
Bundled Scripts &
Skill Refinement &
Doc--Code Consistency &
Repair \& Preservation \\
\midrule
SkillsBench & $\checkmark$ & $\times$ & $\times$ & $\times$ \\
SWE-Skills-Bench & $\times$ & $\times$ & $\times$ & $\times$ \\
SkillLearnBench & $\times$ & $\checkmark$ & $\times$ & $\times$ \\
SkillGenBench & $\checkmark$ & $\checkmark$ & $\times$ & $\times$ \\
\addlinespace[1pt]
\textbf{SkillScriptBench} & $\checkmark$ & $\checkmark$ & $\checkmark$ & $\checkmark$ \\
\bottomrule
\end{tabular*}
\end{table}
\endgroup

\subsection{Task formulation}
\label{sec:task}

Each \system{} task provides an existing executable Agent Skill package $S$ and
a natural-language request $q$ describing the required functionality or a failure
to be resolved. Revision targets the natural-language instructions and behavioral
specifications in \texttt{SKILL.md}, the implementation code of bundled executable
scripts, or both. A method returns a candidate package $S'$ whose executable
behavior satisfies $q$ and whose documentation consistently describes that
behavior, while preserving behavior outside the requested scope.

\subsection{Benchmark construction}
\label{sec:benchmark-construction}

We survey 35,554 GitHub-hosted Skill roots and identify 4,981 containing
source files under \texttt{scripts/} or \texttt{script/}. We select packages
that permit redistribution and whose behavior can be checked using repository
tests or task-specific executable tests. We record the source repository and
commit for each selected package.

\paragraph{In-the-Wild Repair.}
We construct 150 repair tasks from 100 packages by introducing script faults
based on fault patterns identified in our GitHub survey. Each task pairs the faulty package with a
maintenance request describing the expected behavior or a failure symptom.
Executable checks distinguish the faulty version from a correct reference
implementation. Tasks from the same package target different maintenance requirements.

\paragraph{Controlled Repair.}
This track tests whether methods repair faulty documentation and scripts while
preserving correct ones. For each of 50 repository-derived packages, we keep
the maintenance request fixed and construct four states: Clean without faults,
Doc with documentation faults only, Script with script faults only, and Joint
with both.
The 200 tasks support within-package comparisons.

\Needspace{6\baselineskip}
\paragraph{Quality control.}
We check package integrity and alignment between maintenance requests and
tested behavior. Reference packages must pass the task-specific checks, while
faulty versions must fail checks for the injected faults~\citep{jia2011mutation}.
For Controlled Repair, we verify that Doc and Script retain the unaffected
artifact and that Joint combines their faults under the same request.
Appendix~\ref{app:benchmark-construction-audits} details these construction checks.

\begin{figure}[!t]
\centering
\setlength{\abovecaptionskip}{4pt}
\includegraphics[width=0.88\linewidth]{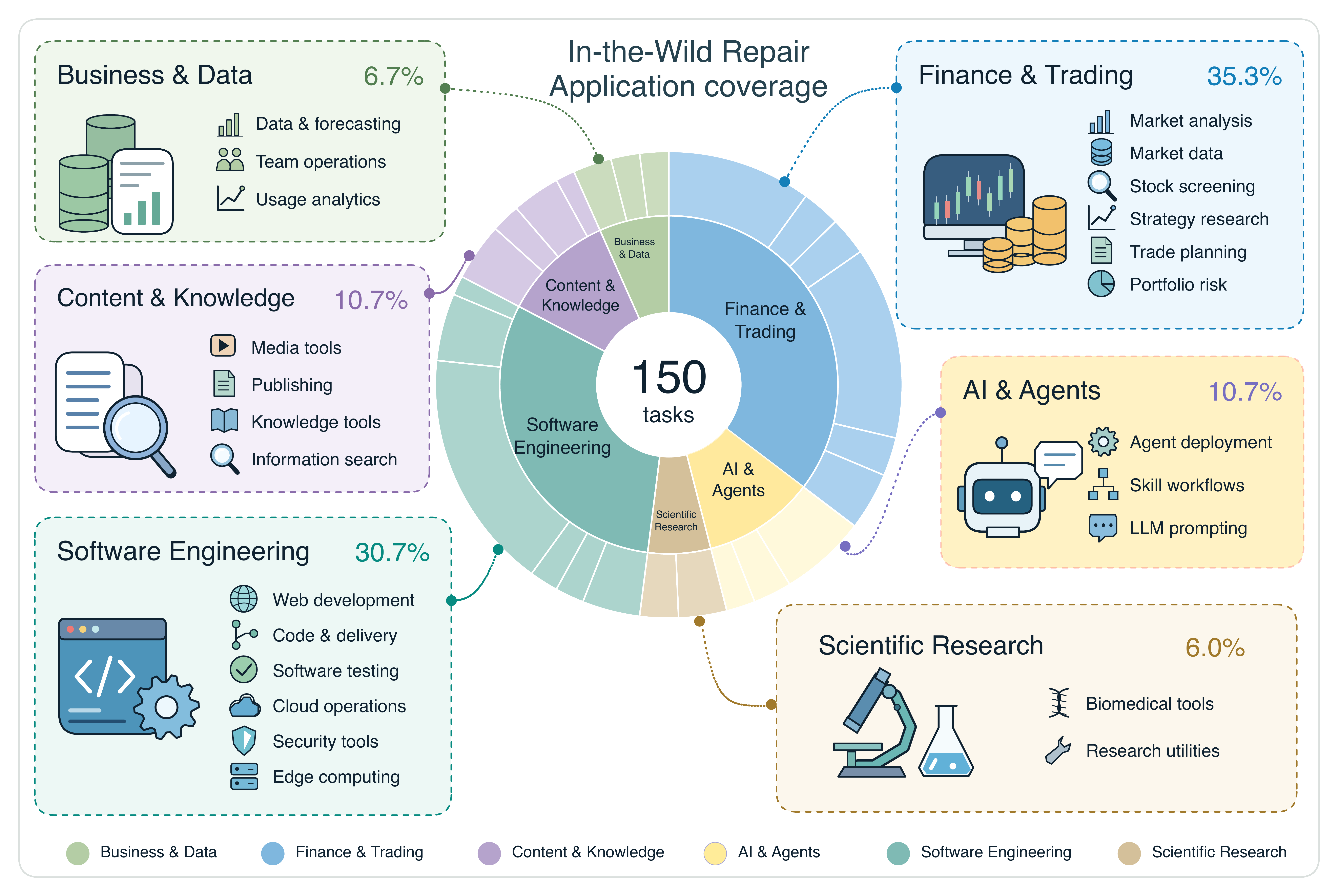}
\caption{Application coverage of In-the-Wild Repair. Inner and outer rings show
six application domains and 24 uses, respectively. Uses are listed clockwise
within each domain.}
\label{fig:benchmark-composition}
\end{figure}

\Needspace{6\baselineskip}
\subsection{Benchmark composition}
\label{sec:benchmark-composition}

Figure~\ref{fig:benchmark-composition} summarizes application coverage;
Appendix Figure~\ref{fig:benchmark-language-scope} compares language and
composite annotation scores across the two tracks. In-the-Wild spans six application domains and
is predominantly Python, whereas
Controlled Repair has a more even distribution across Python, JavaScript, and Shell,
alongside mixed-language packages.

We annotate tasks by maintenance requirement and programming language, and
group In-the-Wild tasks by request localization. Controlled Repair comprises
three diagnostic partitions: Single-Target Repair targets individual script
locations; Multi-Target Repair requires coordinated edits across multiple script
locations; Cross-Call Repair targets caller--helper dependencies.
Appendix~\ref{app:benchmark-construction-audits} details task annotations and composition.

\subsection{Evaluation framework}
\label{sec:benchmark-evaluation}

We evaluate functional correctness and documentation--script consistency.
Behavioral tests check the requested
functionality and preservation of correct behavior, while documentation-driven
tests execute task-relevant invocations and workflows from \texttt{SKILL.md}
against the revised scripts. Success requires passing all applicable tests
in both categories.

\Needspace{6\baselineskip}
Behavioral tests check task-specific outputs, side effects, and argument
propagation. Documentation-driven tests use controlled inputs and simulate
external services where needed. For Clean and Doc-fault tasks, preservation
checks verify that the revised scripts retain the package's existing correct
functionality. All methods are evaluated against the same task-specific
criteria. Appendix~\ref{app:evaluation-details} details the execution protocols
and evaluator calibration.

\FloatBarrier
\Needspace{9\baselineskip}
\section{AST-Guided Skill Revision}
\label{sec:method}

\begin{figure}[!htbp]
\centering
\includegraphics{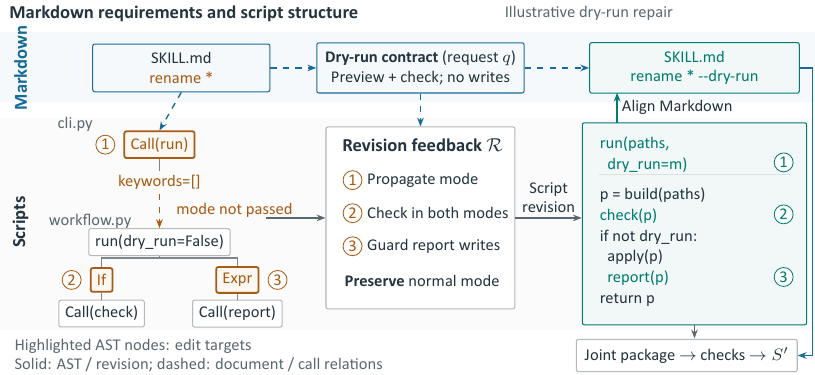}
\caption{AST-Guided Skill Revision on a dry-run example.
Numbered issues (1--3) map through revision feedback $\mathcal R$ to
corresponding script edits, enabling preview generation and conflict checking
without renaming files or writing reports in dry-run mode. The \texttt{SKILL.md}
command is updated with \texttt{--dry-run} to match the repaired implementation.}
\label{fig:full-coverage-method}
\end{figure}

AST-Guided Skill Revision coordinates Markdown--script repair by linking
maintenance requirements to executable structure. We assess documentation and
implementation separately against the same request, identifying discrepancies
to repair and correct behavior to preserve. Given an initial revision
$\widetilde S=M(q,S)$ from a base method $M$, we analyze the original package
$S$ and $\widetilde S$ to derive repair targets and preservation constraints.
These guide subsequent script edits, after which we align documented usage
with $q$ and the revised implementation to produce the final package $S'$
(Figure~\ref{fig:full-coverage-method}).

\subsection{Requirement--Structure Grounding}

The LLM interprets the behavior required by $q$ and identifies related claims
in Markdown. It links these requirements and claims to their applicability
conditions, supporting passages, and relevant implementation sites.
Language-native AST/CST analysis binds the proposed sites to uniquely matched
syntax nodes. Caller--helper relationships connect these nodes to related
implementation sites, providing context for requirements that span function
boundaries.

\subsection{Repair Targets and Preservation Constraints}

\looseness=-1
For each grounded requirement, the LLM assesses the implementation and
documentation separately under its applicability conditions. We map assessed implementation
discrepancies to resolved, editable syntax nodes, forming repair targets that
define the scope of subsequent script edits.
Behavior assessed as satisfying the requirement supplies
preservation constraints, including when the corresponding documentation
requires correction. Related callers and helpers provide supporting context
without necessarily becoming edit targets. We assess both $S$ and
$\widetilde S$, retaining requirement-linked symbols and call context from $S$
when the initial revision removes them. Together, these assessments and
structural bindings form the revision feedback $\mathcal R=G(q,S,\widetilde S)$
supplied to the script editor.

\subsection{Coordinated Markdown--Script Revision}

\looseness=-1
The revision model uses the repair targets, caller--helper context, and
preservation constraints in $\mathcal R$ to generate node-level script
replacements for each package version. Edits leave source text outside the
selected spans unchanged and are accepted only if the resulting scripts pass
syntax and applicable interface-preservation checks. Assembly uses
$\widetilde S$ as its base, retaining its validated edits and incorporating
compatible edits from $S$ through unique source matching.
The model then aligns documented invocations and operational instructions with
the maintenance request $q$ and the resulting scripts, correcting identified
inconsistencies through localized Markdown edits supported by script evidence.
Appendices~\ref{app:structural-backends} and~\ref{app:revision-algorithm} detail
the feedback representation, edit validation, and package assembly rules.

\subsection{Integration with CoEvoSkills}
\label{sec:coevo-integration}

We adapt CoEvoSkills~\citep{zhang2026coevoskills} to maintenance-request-driven
revision of executable Skill packages. An independent surrogate verifier
assesses the package against the request and produces requirement-linked checks
and a diagnosis. A failure reported by the verifier or its executed checks
triggers the generator to revise \texttt{SKILL.md} and scripts using the
diagnosis and check results. To obtain CoEvoSkills + AST, we use the returned
package as the initial revision $\widetilde S$ and apply AST-Guided Skill
Revision to $(q,S,\widetilde S)$.

\Needspace{36\baselineskip}
\section{Experiments}
\label{sec:experiments}
\label{sec:companion-evaluation}
\makeatletter
\newcommand{\MainResultsASTHighlight}{%
  \noalign{\begingroup\color{black!7}%
    \hrule height\dimexpr\ht\@arstrutbox+\dp\@arstrutbox\relax
    \vskip-\dimexpr\ht\@arstrutbox+\dp\@arstrutbox\relax
  \endgroup}}
\makeatother
\newcommand{\MainResultsFiveMethodsRows}{%
\multicolumn{16}{@{}l}{\textbf{GPT-5.5}} \\[1pt]
MD-only & 0.0 & 0.0 & 0.0 & 26.0 & 42.0 & 12.0 & 0.0 & 0.0 & 0.0 & 0.0 & 0.0 & 0.0 & 4.3 & 7.0 & 2.0 \\
Raw & 53.1 & 76.7 & 27.3 & 30.7 & 44.0 & 18.0 & 45.3 & 48.0 & 42.0 & 48.7 & 52.0 & 46.0 & 47.3 & 62.3 & 31.3 \\
\MainResultsASTHighlight
Raw+AST & 69.8 & 88.0 & 49.3 & \textbf{72.0} & \textbf{88.0} & \textbf{52.0} & 74.7 & 82.0 & 66.0 & 70.7 & 80.0 & 60.0 & 71.1 & 85.7 & 54.3 \\
CoEvo & 50.0 & 72.7 & 26.0 & 31.3 & 50.0 & 16.0 & 43.3 & 70.0 & 18.0 & 52.0 & 80.0 & 20.0 & 46.1 & 69.7 & 22.0 \\
\MainResultsASTHighlight
CoEvo+AST & \textbf{74.7} & \textbf{90.7} & \textbf{54.7} & 64.7 & 82.0 & 48.0 & \textbf{80.7} & \textbf{92.0} & \textbf{70.0} & \textbf{79.3} & \textbf{90.0} & \textbf{68.0} & \textbf{74.8} & \textbf{89.3} & \textbf{58.3} \\
\addlinespace[1.2pt]
\midrule
\multicolumn{16}{@{}l}{\textbf{GPT-5.6-Sol}} \\[1pt]
MD-only & 0.0 & 0.0 & 0.0 & 46.7 & 60.0 & 28.0 & 0.0 & 0.0 & 0.0 & 0.0 & 0.0 & 0.0 & 7.8 & 10.0 & 4.7 \\
Raw & 68.2 & 84.7 & 50.0 & 38.7 & 60.0 & 22.0 & 64.7 & 74.0 & 52.0 & 62.7 & 72.0 & 50.0 & 61.8 & 76.7 & 45.7 \\
\MainResultsASTHighlight
Raw+AST & 80.7 & 90.7 & \textbf{67.3} & \textbf{75.3} & \textbf{92.0} & \textbf{52.0} & 83.3 & \textbf{90.0} & 72.0 & 81.3 & \textbf{90.0} & 68.0 & 80.3 & \textbf{90.7} & 65.7 \\
CoEvo & 64.2 & 80.0 & 41.3 & 39.3 & 64.0 & 12.0 & 74.0 & 86.0 & 60.0 & 70.0 & 84.0 & 50.0 & 62.7 & 79.0 & 41.0 \\
\MainResultsASTHighlight
CoEvo+AST & \textbf{81.6} & \textbf{93.3} & 66.0 & 70.0 & 86.0 & \textbf{52.0} & \textbf{86.7} & \textbf{90.0} & \textbf{82.0} & \textbf{86.0} & 88.0 & \textbf{82.0} & \textbf{81.2} & \textbf{90.7} & \textbf{69.0} \\
\addlinespace[1.2pt]
\midrule
\multicolumn{16}{@{}l}{\textbf{Claude Opus 5}} \\[1pt]
MD-only & 0.0 & 0.0 & 0.0 & 23.3 & 28.0 & 20.0 & 0.0 & 0.0 & 0.0 & 0.0 & 0.0 & 0.0 & 3.9 & 4.7 & 3.3 \\
Raw & 66.4 & 77.3 & 55.3 & 26.7 & 32.0 & 22.0 & 50.7 & 56.0 & 46.0 & 57.3 & 60.0 & 54.0 & 55.7 & 63.3 & 48.0 \\
\MainResultsASTHighlight
Raw+AST & \textbf{80.9} & \textbf{90.7} & \textbf{68.7} & 66.0 & \textbf{82.0} & 48.0 & \textbf{78.7} & 82.0 & \textbf{76.0} & \textbf{83.3} & 86.0 & \textbf{80.0} & \textbf{78.4} & \textbf{87.0} & \textbf{68.3} \\
CoEvo & 44.4 & 50.7 & 36.7 & 8.7 & 10.0 & 8.0 & 46.0 & 56.0 & 36.0 & 52.7 & 60.0 & 46.0 & 40.1 & 46.3 & 33.3 \\
\MainResultsASTHighlight
CoEvo+AST & 74.0 & 78.7 & \textbf{68.7} & \textbf{72.0} & 80.0 & \textbf{58.0} & 76.7 & \textbf{90.0} & 66.0 & 76.7 & \textbf{90.0} & 64.0 & 74.6 & 82.7 & 65.7 \\
\addlinespace[1.2pt]
\midrule
\multicolumn{16}{@{}l}{\textbf{DeepSeek V4 Pro}} \\[1pt]
MD-only & 0.0 & 0.0 & 0.0 & 12.0 & 12.0 & 12.0 & 0.0 & 0.0 & 0.0 & 0.0 & 0.0 & 0.0 & 2.0 & 2.0 & 2.0 \\
Raw & 61.1 & 64.0 & 57.3 & 13.3 & 14.0 & 12.0 & 66.0 & 68.0 & 64.0 & 63.3 & 64.0 & 62.0 & 54.3 & 56.3 & 51.7 \\
\MainResultsASTHighlight
Raw+AST & \textbf{78.2} & \textbf{81.3} & \textbf{74.7} & 64.0 & 76.0 & 48.0 & 82.0 & 84.0 & 80.0 & 80.0 & 82.0 & \textbf{78.0} & \textbf{76.8} & \textbf{81.0} & \textbf{71.7} \\
CoEvo & 50.7 & 56.7 & 44.7 & 13.3 & 20.0 & 8.0 & 54.0 & 58.0 & 50.0 & 58.7 & 66.0 & 50.0 & 46.3 & 52.3 & 40.3 \\
\MainResultsASTHighlight
CoEvo+AST & 70.9 & 75.3 & 66.7 & \textbf{74.7} & \textbf{90.0} & \textbf{58.0} & \textbf{84.0} & \textbf{86.0} & \textbf{82.0} & \textbf{81.3} & \textbf{84.0} & \textbf{78.0} & 75.4 & \textbf{81.0} & 69.7 \\
}

\begin{table}[H]
\caption{Repair success on \system{} across four LLM backbones (\%).}
\label{tab:cross-model-results}
\centering
\small
\setlength{\tabcolsep}{1.5pt}
\renewcommand{\arraystretch}{0.97}
\begin{tabular*}{\linewidth}{@{\extracolsep{\fill}}l@{\hspace{6pt}}rrr@{\hspace{7pt}}rrr@{\hspace{5pt}}rrr@{\hspace{5pt}}rrr@{\hspace{7pt}}rrr@{}}
\toprule
& \multicolumn{3}{c}{\multirow{2}{*}{In-the-Wild}}
& \multicolumn{9}{c}{Controlled Repair}
& \multicolumn{3}{c}{\multirow{2}{*}{Overall}} \\
\cmidrule(lr){5-13}
& & & & \multicolumn{3}{c}{Doc.} & \multicolumn{3}{c}{Script}
& \multicolumn{3}{c}{Joint} & & & \\
\cmidrule(lr){2-4}\cmidrule(lr){5-7}\cmidrule(lr){8-10}\cmidrule(lr){11-13}\cmidrule(lr){14-16}
Method & Avg & P@3 & $\mathrm{Hit}^{3}$ & Avg & P@3 & $\mathrm{Hit}^{3}$ & Avg & P@3 & $\mathrm{Hit}^{3}$ & Avg & P@3 & $\mathrm{Hit}^{3}$ & Avg & P@3 & $\mathrm{Hit}^{3}$ \\
\midrule
\MainResultsFiveMethodsRows
\bottomrule
\end{tabular*}
\par\smallskip
\begin{minipage}{\linewidth}
\small
MD-only, Raw, and CoEvo denote Markdown-only, Raw Package, and CoEvoSkills.
Avg reports the mean success rate over three runs; P@3 and $\mathrm{Hit}^{3}$
report the fractions of tasks successful in at least one and all three runs, respectively.
Doc., Script, and Joint denote documentation, script, and joint faults.
Overall pools 150 faulty-package tasks from each track.
Figure~\ref{fig:scope-revision-analysis} reports preservation on the 50 Clean tasks.
Grey rows mark +AST methods; bold indicates column maxima per model, including ties.
\end{minipage}
\end{table}

\subsection{Experimental Setup}

\paragraph{Models and benchmark scope.}
We evaluate all methods with GPT-5.5, GPT-5.6-Sol, Claude Opus 5, and DeepSeek
V4 Pro on all 350 SkillScriptBench tasks: 150 In-the-Wild repair cases
and 200 Controlled Repair tasks.
Evaluation follows the protocol in Section~\ref{sec:benchmark-evaluation}.

\paragraph{Compared methods.}
We compare five settings, each given the task-specific maintenance request.
\emph{Markdown-only} receives Markdown inputs and
edits only \texttt{SKILL.md}. \emph{Raw Package} and \emph{CoEvoSkills}
receive the full package and can edit \texttt{SKILL.md} and scripts;
CoEvoSkills uses surrogate-verifier feedback to guide package revision,
following the adaptation in Section~\ref{sec:coevo-integration}.
We augment Raw Package and CoEvoSkills with AST-Guided Skill Revision for
coordinated Markdown--script repair, obtaining \emph{Raw Package + AST}
and \emph{CoEvoSkills + AST}, respectively.
Appendix~\ref{app:artifactstate200-reporting} details the model configurations,
revision protocols, and prompt templates.

\Needspace{11\baselineskip}
\paragraph{Repetitions and metrics.}
We report three runs for each model--method--task combination.
Avg (Pass@1) is the mean success rate across runs; Pass@3 (P@3) and
$\mathrm{Hit}^{3}$ are the fractions of tasks solved in at least one run and
all three runs, respectively~\citep{chen2021codex,yao2024taubench}.
Table~\ref{tab:cross-model-results} reports repair success, with Overall pooling
the 300 faulty-package tasks. Figure~\ref{fig:scope-revision-analysis} summarizes
preservation on the 50 Clean tasks; Appendix Table~\ref{tab:clean-preservation}
gives per-model results. Cross-model means weight the four
models equally. Appendix Table~\ref{tab:pass-at-three} additionally reports P@3
on Clean tasks and the full 350-task benchmark.
We report paired source-package bootstrap confidence intervals~\citep{davison1997bootstrap} for Overall Avg
gains in Appendix Table~\ref{tab:paired-gain-intervals}.

\Needspace{7\baselineskip}
\subsection{Main Results}

\textbf{Full-package baselines achieve broader repair coverage, but documentation repair
remains a weakness.} Averaged across four backbones, Raw Package and CoEvoSkills
achieve Overall Avg scores of 54.8\% and 48.8\%, respectively. Both baselines
repair script faults, joint faults, and In-the-Wild cases. However, their
documentation-repair Avg scores are
27.3\% and 23.2\%, respectively, compared with 27.0\% for Markdown-only.

\Needspace{6\baselineskip}
\textbf{AST-guided revision improves repair success across all four backbones.}
Raw Package + AST and CoEvoSkills + AST achieve Overall Avg
scores of 76.7\% and 76.5\%, improving over their respective baselines by 21.9
and 27.7 pp. Their P@3 reaches 86.1\% and 85.9\%, while
$\mathrm{Hit}^{3}$ reaches 65.0\% and 65.7\%, respectively. All three metrics
improve for both generators on every backbone.

\FloatBarrier
\Needspace{14\baselineskip}
\section{Analysis}
\label{sec:analysis}
\label{sec:results}

We analyze package revision along two dimensions: the balance between fault
correction and behavior preservation, and the variation in revision gains
across task characteristics.

\subsection{RQ1: What does AST-guided revision add beyond full-package access?}
\label{sec:analysis-artifact-state}
\label{sec:diagnosis}
\label{sec:baseline-study}

\begin{figure}[!htbp]
\centering
\includegraphics[width=\linewidth]{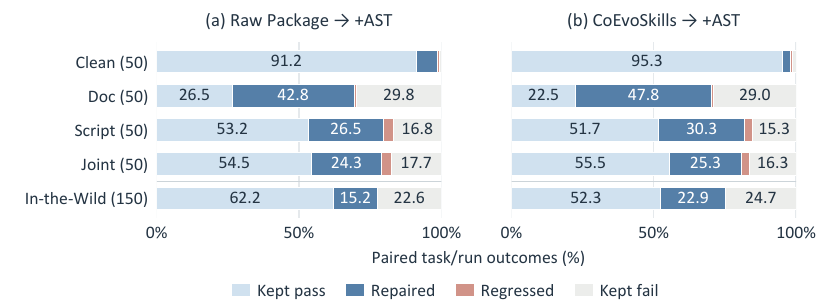}
\caption{AST-guided revision improves repair and preservation.
Paired outcomes for (a) Raw Package and (b) CoEvoSkills before and after
AST-guided revision, pooled across four backbones and three runs.
Parentheses indicate task counts.}
\label{fig:scope-revision-analysis}
\end{figure}

\looseness=-1
Broader repair coverage does not imply better preservation of correct behavior.
As reported in Appendix Table~\ref{tab:clean-preservation}, Raw Package and
CoEvoSkills achieve lower Avg on Clean tasks than Markdown-only across all four
backbones, with mean gaps of 7.8 and 3.7 percentage points, respectively.
These results highlight the need to distinguish what must change from what
should remain intact during package revision.

AST-guided revision improves fault correction and behavior preservation together.
By linking Markdown requirements to script structure, its feedback provides
both repair targets and preservation constraints.
As shown in Figure~\ref{fig:scope-revision-analysis}, AST-guided revision
corrects substantially more failures than it introduces. Applied to Raw Package
on documentation-fault tasks, it changes 42.8\% of task/run pairs from failure
to success, whereas only 0.8\% change from success to failure. Corrections also outnumber
regressions on script and joint faults for both generators. On Clean tasks,
mean $\mathrm{Hit}^{3}$ rises from 85.5\% to 96.5\% for Raw Package and from 93.0\% to 96.5\%
for CoEvoSkills, indicating more consistent preservation across all three runs.

In the ablation study on Controlled Repair with GPT-5.6-Sol,
\emph{w/o Structural Feedback} and \emph{w/o Call Context} retain Markdown
alignment and achieve 69.2\% and 67.3\% Avg, respectively, compared with 84.8\%
for AST-Guided Skill Revision. The full configuration outperforms both variants
across all three fault states and in Clean-state preservation, with the largest
margins on Joint tasks. Appendix~\ref{app:feedback-variants} provides configuration
details and state-wise results.

\FloatBarrier

\subsection{RQ2: How do revision gains vary across task characteristics?}
\label{sec:analysis-models}
\label{sec:analysis-partitions}
\label{sec:analysis-fault-family}
\label{sec:analysis-language}
\label{sec:analysis-request-mode}

We compare Raw Package with Raw Package + AST on faulty-package tasks, examining
structural partitions in Controlled Repair and repair categories, programming
languages, and request localization in In-the-Wild.

\begin{figure}[!htbp]
\centering
\setlength{\abovecaptionskip}{4pt}
\includegraphics[width=\linewidth]{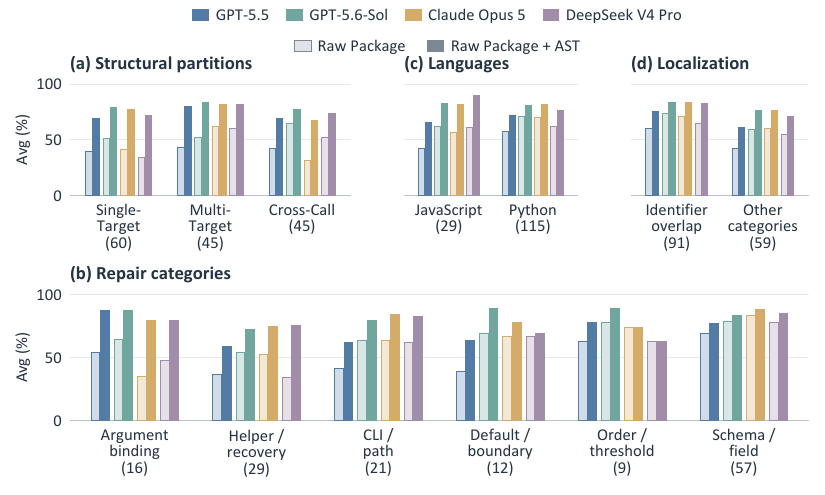}
\caption{Revision gains vary across task characteristics.
Panel (a) uses faulty-package tasks from Controlled Repair; panels (b--d)
use In-the-Wild. Light/dark bars show Raw Package/Raw Package + AST
Avg over three runs for each model. Parentheses give task counts;
complete repair-category results appear in Appendix Figure~\ref{fig:category-heatmap}.}
\label{fig:fault-family-gains}
\end{figure}

\Needspace{4\baselineskip}
\paragraph{Structural partitions.}
AST-guided revision improves both single-target and coordinated repairs across
all four backbones. As shown in Figure~\ref{fig:fault-family-gains}(a), mean Avg
increases by 32.8, 27.6, and 24.1 percentage points on Single-Target,
Multi-Target, and Cross-Call, respectively. Multi-Target reaches the highest
post-revision Avg of 81.9\% and outperforms Single-Target across all four
backbones, demonstrating that the benefits extend beyond isolated edits.
Cross-Call improves to 71.9\% but remains the least successful partition.

\Needspace{9\baselineskip}
\paragraph{Repair categories.}
AST-guided revision yields its largest gains on argument-binding and
helper/recovery repairs. As shown in Figure~\ref{fig:fault-family-gains}(b),
their Avg rises from 50.5\% to 83.3\% and from 44.5\% to 70.4\%, respectively.
Argument-binding repairs thus reach nearly the same success rate as
schema/field mapping, which improves from a higher baseline of 77.3\% to
83.5\%.
Avg increases from 27.8\% to 30.6\% on the three
branch/route tasks and remains at 61.1\% on \mbox{the three multi-file pipeline tasks.}

\Needspace{8\baselineskip}
\paragraph{Programming languages.}
AST-guided revision produces larger gains on JavaScript than Python across all
four backbones. Figure~\ref{fig:fault-family-gains}(c) shows that cross-model
mean Avg rises from 55.5\% to 79.9\% for JavaScript and from 64.9\% to 78.0\%
for Python. The two language groups thus reach similar post-revision success
rates despite JavaScript's substantially lower baseline.

\paragraph{Request localization.}
AST-guided revision reduces the performance gap across request-localization
categories. As shown in Figure~\ref{fig:fault-family-gains}(d), Avg improves by
13.8 percentage points for identifier-overlap tasks and 17.2 for the remaining
tasks, narrowing the cross-model mean gap from 13.4 to 10.0 points.

\section{Related Work}
\label{sec:related}

\paragraph{Agent Skill Self-Evolution.}
SkillOpt and SkillAudit focus on revising skill documents, using bounded textual edits
and paired-trajectory feedback, respectively~\citep{yang2026skillopt,gao2026skillaudit}.
CoEvoSkills and SkillEvolver support broader package-level evolution, jointly constructing
and refining natural-language instructions and executable
code~\citep{zhang2026coevoskills,zhang2026skillevolver}. However, downstream task success alone
does not separately assess documentation repair, script repair,
documentation--script consistency, and preservation of correct behavior.

\paragraph{Benchmarks for Agent Skills.}
SkillsBench and SWE-Skills-Bench evaluate the utility of supplied skills across diverse agent tasks
and repository-based software engineering, respectively~\citep{li2026skillsbench,han2026sweskillsbench}.
SkillLearnBench evaluates continual skill generation through specification quality, execution
trajectories, and task outcomes, while SkillGenBench compares pipelines that distill repositories
and documents into reusable skills~\citep{zhong2026skilllearnbench,zhou2026skillgenbench}.
However, these benchmarks do not systematically evaluate the repair of existing Skill packages
comprising both Markdown instructions and executable scripts.

\Needspace{7\baselineskip}
\paragraph{Program Repair.}
Structure-aware program repair uses program representations to guide code changes.
Astor applies AST-level transformations to Java programs, Getafix learns hierarchical fix patterns
from past edits, and TBar applies repair templates at localized fault
sites~\citep{martinez2016astor,bader2019getafix,liu2019tbar}. LLM-based approaches such as
Agentless organize repository repair into localization, patch generation, and
validation~\citep{xia2024agentless}. Repository-level benchmarks such as
SWE-bench evaluate patches against real issue descriptions and executable
tests~\citep{jimenez2024swebench}. Complementary work updates natural-language comments
to reflect changes in the associated code~\citep{panthaplackel2020learning}.
These settings do not explicitly address joint revision
of Markdown instructions and executable scripts to restore their consistency.\looseness=-1

\section{Conclusion}
\label{sec:conclusion}

We introduced \system{}, a 350-task benchmark for self-evolution of executable
Agent Skill packages. It combines In-the-Wild Repair, built from a scan of over
35,000 public Skill roots, with Controlled Repair across matched clean,
documentation-fault, script-fault, and joint-fault states. Across four LLM backbones,
full-package baselines broaden repair coverage but do not consistently outperform
Markdown-only revision in documentation repair or clean-package preservation. We proposed
AST-Guided Skill Revision to coordinate Markdown and script revisions through
requirement-grounded structural feedback. Applied to Raw Package and CoEvoSkills,
it improves both repair and preservation while increasing mean all-three-run
success ($\mathrm{Hit}^{3}$) across all four artifact states.
Evaluation of executable Skill evolution should distinguish fault repair from
preservation of correct package behavior.

\FloatBarrier
\bibliography{references}
\bibliographystyle{iclr2027_conference}

\appendix
\raggedbottom
\section{Benchmark, method, and evaluation details}
\label{app:protocol}
\setlength{\intextsep}{7pt plus 2pt minus 2pt}
\makeatletter
\setlength{\@fptop}{0pt}
\setlength{\@fpsep}{14pt}
\setlength{\@fpbot}{0pt plus 1fil}
\makeatother
\newcommand{\CurrentRunMDGptFiveFiveRow}{& Markdown-only & 0 & 0 & 0 & 66 & 61 & 62 \\}
\newcommand{\CurrentRunCoevoGptFiveFiveRow}{& CoEvoSkills & 90 & 71 & 64 & 116 & 111 & 106 \\}
\newcommand{\CurrentRunCoevoASTGptFiveFiveRow}{& CoEvoSkills + AST & 118 & 112 & 106 & 164 & 166 & 152 \\}
\newcommand{\CurrentRunRawGptFiveFiveRow}{& Raw Package & 72 & 86 & 81 & 105 & 106 & 113 \\}
\newcommand{\CurrentRunRawASTGptFiveFiveRow}{& Raw Package + AST & 104 & 109 & 101 & 160 & 156 & 157 \\}
\newcommand{\CurrentRunMDGptFiveSixRow}{& Markdown-only & 0 & 0 & 0 & 69 & 77 & 74 \\}
\newcommand{\CurrentRunCoevoGptFiveSixRow}{& CoEvoSkills & 80 & 99 & 110 & 136 & 139 & 141 \\}
\newcommand{\CurrentRunCoevoASTGptFiveSixRow}{& CoEvoSkills + AST & 119 & 121 & 127 & 170 & 173 & 167 \\}
\newcommand{\CurrentRunRawGptFiveSixRow}{& Raw Package & 102 & 105 & 100 & 132 & 131 & 119 \\}
\newcommand{\CurrentRunRawASTGptFiveSixRow}{& Raw Package + AST & 120 & 121 & 122 & 175 & 168 & 166 \\}
\newcommand{\CurrentRunMDOpusFiveRow}{& Markdown-only & 0 & 0 & 0 & 62 & 61 & 60 \\}
\newcommand{\CurrentRunCoevoOpusFiveRow}{& CoEvoSkills & 66 & 67 & 67 & 98 & 107 & 101 \\}
\newcommand{\CurrentRunCoevoASTOpusFiveRow}{& CoEvoSkills + AST & 109 & 113 & 111 & 154 & 171 & 161 \\}
\newcommand{\CurrentRunRawOpusFiveRow}{& Raw Package & 104 & 95 & 100 & 116 & 113 & 110 \\}
\newcommand{\CurrentRunRawASTOpusFiveRow}{& Raw Package + AST & 123 & 117 & 124 & 166 & 162 & 160 \\}
\newcommand{\CurrentRunMDDeepSeekRow}{& Markdown-only & 0 & 0 & 0 & 56 & 56 & 56 \\}
\newcommand{\CurrentRunCoevoDeepSeekRow}{& CoEvoSkills & 76 & 76 & 76 & 109 & 114 & 113 \\}
\newcommand{\CurrentRunCoevoASTDeepSeekRow}{& CoEvoSkills + AST & 106 & 106 & 107 & 172 & 166 & 172 \\}
\newcommand{\CurrentRunRawDeepSeekRow}{& Raw Package & 89 & 93 & 93 & 118 & 119 & 121 \\}
\newcommand{\CurrentRunRawASTDeepSeekRow}{& Raw Package + AST & 114 & 121 & 117 & 166 & 161 & 161 \\}

\subsection{Benchmark composition and construction checks}
\label{app:benchmark-construction-audits}

We detail the construction and validation of both tracks, followed by the
annotations used to describe their composition. Table~\ref{tab:tracks}
summarizes their sizes, language coverage, and task construction.

\begin{table}[!htbp]
\caption{Benchmark sizes, language coverage, and task construction.
Language counts refer to repair cases for In-the-Wild and canonical base
packages for Controlled Repair. TypeScript and mixed-language packages are
reported separately here and grouped as Others in Figure~\ref{fig:benchmark-language-scope}.}
\label{tab:tracks}
\centering
\small
\setlength{\tabcolsep}{3.5pt}
\begin{tabular}{@{}>{\raggedright\arraybackslash}p{0.17\linewidth}>{\raggedright\arraybackslash}p{0.15\linewidth}>{\raggedright\arraybackslash}p{0.27\linewidth}>{\raggedright\arraybackslash}p{0.30\linewidth}@{}}
\toprule
Track & Size & Language profile & Task construction \\
\midrule
In-the-Wild & 150 cases & 115 Python; 29 JS; 3 TS; 3 Shell & Script-fault injection \\
Controlled Repair & $50\times4=200$ tasks & 18 Python; 14 JS; 14 Shell; 2 TS; 2 mixed & Matched Clean, Doc, Script, and Joint states \\
\bottomrule
\end{tabular}
\end{table}

\paragraph{In-the-Wild sources and case construction.}
We select packages with licenses permitting
redistribution, traceable source revisions, and task-relevant behavior that can
be tested deterministically under controlled conditions. Repair cases are
validated using repository tests or contract-derived probes that accept
reference implementations and reject the corresponding faulty variants.

\begin{figure}[H]
\centering
\includegraphics[width=\linewidth]{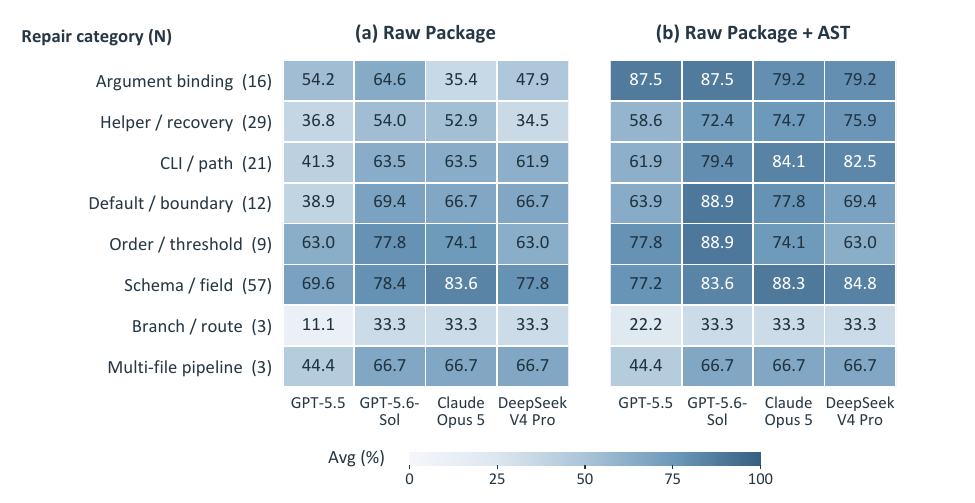}
\caption{AST-guided revision improves argument-binding and helper/recovery
repairs across all four backbones. Cells show In-the-Wild Avg (\%) over
three runs; $N$ is the number of repair tasks. Both panels share a
0--100\% color scale.}
\label{fig:category-heatmap}
\end{figure}

We construct repair cases by introducing source-level faults into repository-derived Skill packages
through parameter substitution, helper-call bypassing, and changes to
default-value handling. Each case pairs the faulty package with a maintenance
request specifying the required behavior or describing a failure symptom.
Cases derived from the same package target distinct behavioral requirements.
The 150 cases span eight repair categories, with their distribution reported
in Figure~\ref{fig:category-heatmap}.
The case registry records source repositories, commits, package identities, and
content hashes. Fault and state labels, reference packages, and benchmark
acceptance checks are reserved for construction and evaluation and excluded
from method inputs.

\paragraph{Controlled Repair bases.}
Controlled Repair uses 50 canonical Skill packages derived from repository
sources. Construction records document their source revisions, licenses, and
injected faults. Each package pairs a reference implementation
with a description of the behavior targeted by its maintenance request.
For Multi-Target and Cross-Call, this behavior is specified in an
operational-contract block in \texttt{SKILL.md}.
Table~\ref{tab:state-construction} summarizes the four states constructed
under the same maintenance request.

\begin{table}[!htb]
\caption{Four matched Controlled Repair states. Unchanged denotes the corresponding
artifact from the canonical package before revision.}
\label{tab:state-construction}
\centering
\small
\setlength{\tabcolsep}{3.5pt}
\begin{tabular*}{\linewidth}{@{\extracolsep{\fill}}llll@{}}
\toprule
State & Markdown & Scripts & Evaluation focus \\
\midrule
Clean & Unchanged & Unchanged & Preservation \\
Doc. & Fault-injected & Unchanged & Documentation repair \\
Script & Unchanged & Fault-injected & Script repair \\
Joint & Fault-injected & Fault-injected & Joint repair \\
\bottomrule
\end{tabular*}
\end{table}

Appendix~\ref{app:joint-repair-case} instantiates these four states for the VCP
screener and follows a Joint task through revision and execution.

\Needspace{7\baselineskip}
\paragraph{Construction checks.}
Static checks verify source metadata.
In each of the
Multi-Target and Cross-Call partitions, all 15 canonical bases and all four
reference-repaired states per base pass the private contract tests during
calibration. Appendix~\ref{app:evaluation-details} reports behavioral-check
calibration for Single-Target and Cross-Call.

\paragraph{Application coverage.}
Application labels follow the source packages' documented purpose.
Counts refer to repair tasks; cases derived
from the same package share its application label.

\begin{figure}[!htb]
\centering
\includegraphics[width=\linewidth]{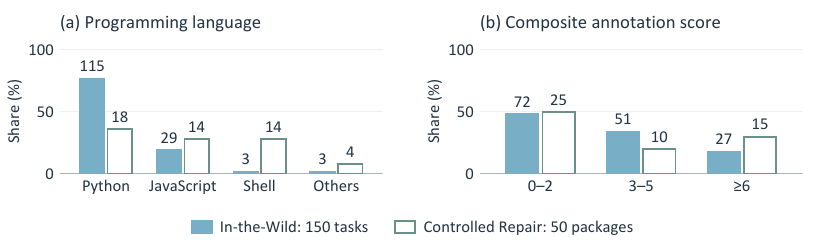}
\caption{Language and composite annotation scores of the two benchmark tracks.
Counts refer to 150 In-the-Wild tasks and 50 Controlled Repair base packages.
Bars show percentages; labels give counts. Others combines TypeScript and
mixed-language packages.}
\label{fig:benchmark-language-scope}
\end{figure}

\paragraph{Language and composite annotations.}
Figure~\ref{fig:benchmark-language-scope} compares the two tracks by programming
language and composite annotation score. The score sums the components in
Table~\ref{tab:composite-annotation-rules} and is grouped as 0--2, 3--5, and
$\geq6$. Table~\ref{tab:artifactstate-partitions} reports the composition of
the three Controlled Repair partitions.

\begin{table}[!htb]
\caption{Components of the composite annotation score. Points are summed
across components; the target-count increments are cumulative.}
\label{tab:composite-annotation-rules}
\centering
\small
\setlength{\tabcolsep}{5pt}
\begin{tabular*}{\linewidth}{@{\extracolsep{\fill}}lll@{}}
\toprule
Component & Condition & Points \\
\midrule
Target count & At least two targets; at least four targets & $+1$; additional $+2$ \\
Target-file count & At least two target files & $+2$ \\
Call-related metadata & Present & $+2$ \\
Request annotation & Symptom-based annotation present & $+1$ \\
Package size & At least ten files & $+1$ \\
Script count & At least three scripts & $+1$ \\
\bottomrule
\end{tabular*}
\par\smallskip
\begin{minipage}{\linewidth}
\footnotesize
For Controlled Repair, the symptom-based request annotation is assigned from construction metadata
or request-text rules. For In-the-Wild, it marks tasks outside the
identifier-overlap category.
\end{minipage}
\end{table}

\begin{table}[!htb]
\caption{Composition of the three Controlled Repair partitions.
Partitions distinguish how repair targets are organized. Language and composite
score counts refer to base packages,
each contributing four tasks with different artifact states.}
\label{tab:artifactstate-partitions}
\centering
\small
\setlength{\tabcolsep}{3pt}
\begin{tabular*}{\linewidth}{@{\extracolsep{\fill}}>{\raggedright\arraybackslash}p{0.15\linewidth}>{\centering\arraybackslash}p{0.12\linewidth}>{\raggedright\arraybackslash}p{0.25\linewidth}rrr@{}}
\toprule
Partition & Bases/tasks & Language profile & Score 0--2 & Score 3--5 & Score $\geq6$ \\
\midrule
Single-Target & 20/80 & 9 JS; 9 Shell; 2 TS & 20 & 0 & 0 \\
Multi-Target & 15/60 & 5 JS; 5 Shell; 3 Python; 2 mixed & 5 & 4 & 6 \\
Cross-Call & 15/60 & 15 Python & 0 & 6 & 9 \\
\bottomrule
\end{tabular*}
\par\smallskip
\begin{minipage}{\linewidth}
\footnotesize
Repair targets: individual script locations (Single-Target), 2--3 script
locations (Multi-Target), and 1--9 mutated call bindings (Cross-Call).
\end{minipage}
\end{table}

\Needspace{5\baselineskip}
\paragraph{Request information and localization.}
Each Controlled Repair base has one maintenance request shared across its
four states. Of these 50 requests, 22 specify behavioral contracts only,
18 describe failure symptoms only, and 10 name an executable path or symbol
associated with the repair target. Requests are checked for replacement-code fragments.

For In-the-Wild, request localization is labeled by matching the request
against package code-file paths, filenames, and extracted identifiers, in that
order. The identifier-overlap category contains 91 tasks; the other categories
comprise 12 path matches, one filename match, and 46 tasks with no detected match.

\subsection{Evaluation details}
\label{app:evaluation-details}

For task $t$, $\mathcal{C}_t$ contains the acceptance checks for required
behavior, documented use, and preservation of correct functionality.
All methods are evaluated with the same task-specific checks. A revised
package $S'$ succeeds when all checks pass:
\[
Y_t(S')=\bigwedge_{c\in\mathcal{C}_t}c(S').
\]
\paragraph{Executable behavior.}
Behavioral tests exercise target functions, script entry points,
caller--helper workflows, or repository test suites. Function-level tests
execute the selected source with its required dependency closure; entry-point
probes import the revised module or launch the revised script. Assertions
check task-relevant return values, emitted files, exit status, output streams,
or arguments passed to downstream calls. Controlled fixtures supply external CLI tools,
HTTP responses, hardware interfaces, and other dependencies as needed.

\paragraph{Documentation-driven execution.}
Task-specific evaluators extract commands and workflows from the revised
\texttt{SKILL.md} and execute them with the revised scripts under predefined
test scenarios. Entry points and arguments come from the documented commands;
working directories and variable bindings come from the surrounding workflow.
Shell parsing preserves quoting and line continuations. Test inputs instantiate
documented placeholders, and controlled fixtures supply external dependencies.
Assertions check exit status, outputs, generated artifacts, and downstream
arguments against the task requirements.

Appendix~\ref{app:joint-repair-case} details a VCP repair case, including
the executed command, test inputs, output assertions, and checks that
referenced package files are present and readable.

\paragraph{Command coverage and verdicts.}
Coverage checks match task-relevant documented invocations to execution records.
Missing invocations, empty extraction results, and unresolved workflow context
are reviewed to identify package defects or evaluation errors. Evaluator and
environment errors are corrected before the unchanged package is re-evaluated.

\paragraph{Preservation.}
For documentation-fault tasks, evaluators execute invocations from both the
revised and reference documentation against the revised scripts. The former
tests the revised usage instructions; the latter tests preservation of
existing behavior. Clean tasks require the revised package to retain both
its behavioral correctness and documented-use correctness.

\paragraph{Execution environment.}
The entry-point probe suite runs as an unprivileged user in a network-disabled
container with a read-only filesystem and a writable temporary directory.
The revised package and probe code are mounted read-only. The standard
container limit is one CPU, 512\,MiB memory, and a 35-second outer timeout;
the browser fixture uses a separate image and a 1\,GiB memory limit.
Other execution backends use task-specific subprocess or
language-runtime boundaries. Execution logs record the package
hash, command, exit status, outputs, and assertion outcomes.

\paragraph{Behavioral-check calibration.}
Test inputs distinguish faulty behavior from the required behavior.
Caller--helper probes use distinct values at different argument positions
so that swapping or reusing an argument produces an observable failure.
For a CLI limit, probes cover an omitted option, zero, and nonzero explicit
values, checking the value in the outgoing API request.

We calibrate behavioral checks using behavior-preserving rewrites and faulty
package variants (Table~\ref{tab:behavioral-calibration}). For each
Single-Target base, the positive control uses an equivalent script rewrite,
such as \texttt{400 <= prev.status} in place of
\texttt{prev.status >= 400}; the negative control appends the correct
fragment as a comment while retaining the script fault. For Cross-Call,
we construct two controls at each of 49 targets, modifying one call argument
in a reference package: the positive variant replaces a variable
\texttt{x} with \texttt{(x,)[0]}, whereas the negative variant substitutes
a different caller variable.

\begin{table}[!htb]
\caption{Behavioral-check calibration on constructed package variants.
Counts give variants with the expected behavioral verdict.}
\label{tab:behavioral-calibration}
\centering
\small
\setlength{\tabcolsep}{4pt}
\begin{tabular*}{\linewidth}{@{\extracolsep{\fill}}lrll@{}}
\toprule
Partition & Bases & Control & Recorded verdict \\
\midrule
Single-Target & 20 & Equivalent script rewrite & 20/20 accepted \\
 & & Comment-only false repair & 20/20 rejected \\
\midrule
Cross-Call & 15 & Equivalent argument expression & 49/49 accepted \\
 & & Single incorrect argument binding & 49/49 rejected \\
\bottomrule
\end{tabular*}
\end{table}

\paragraph{Representative acceptance checks.}
Table~\ref{tab:acceptance-coverage} connects package requirements to
executable checks, with recorded controls and the VCP revisions illustrating
how the checks distinguish valid repairs from faulty outputs.

\begin{table}[!htb]
\caption{Representative acceptance checks and recorded discriminating evidence.
The script controls are summarized in Table~\ref{tab:behavioral-calibration};
the VCP revisions are detailed in Appendix~\ref{app:joint-repair-case}.}
\label{tab:acceptance-coverage}
\centering
\small
\setlength{\tabcolsep}{4pt}
\renewcommand{\arraystretch}{1.12}
\begin{tabular}{@{}>{\raggedright\arraybackslash}p{.18\linewidth}>{\raggedright\arraybackslash}p{.31\linewidth}>{\raggedright\arraybackslash}p{.46\linewidth}@{}}
\toprule
Requirement & Executable check & Discriminating evidence \\
\midrule
Script behavior & Execute target code and downstream argument bindings. &
Accept equivalent rewrites; reject comment-only repairs and incorrect bindings. \\
\addlinespace
Documented entry points & Execute the entry point and options specified by
the document. &
Accept a renamed vault entry point with matching documentation;
reject a documented command with an unsupported option. \\
\addlinespace
Task outputs & Check VCP detections, requested scan settings, and forward
evaluation windows. &
Raw Package misses expected detections despite exiting successfully;
Raw Package + AST passes both contracting price scales and the non-contracting input. \\
\addlinespace
Required artifacts & Inspect created vault directories, document contents,
and verification reports against the fixture state. &
Accept helper extraction and equivalent report wording;
reject missing artifacts and false healthy-status reports. \\
\addlinespace
Existing functionality & Check the new vault reference together with
pre-existing configuration contents. &
Accept the reference control; reject a variant that adds the required
reference but erases unrelated configuration. \\
\bottomrule
\end{tabular}
\end{table}

\FloatBarrier
\subsection{Requirement feedback and structural binding}
\label{app:structural-backends}

\paragraph{Requirement records.}
Requirement records connect requested behavior to source evidence,
implementation assessments, and editable locations in both $S$ and
$\widetilde S$ (Table~\ref{tab:structural-backends}).
The LLM reads $q$, \texttt{SKILL.md}, and the supported scripts in $S$ and
$\widetilde S$ to extract behavioral requirements, applicability conditions,
and supporting passages. Request passages
support explicit requirements and inferences from reported problems;
Markdown passages support document claims assessed against the requested
behavior. Markdown alignment checks the same requested behavior
against the assembled scripts, as described in
Appendix~\ref{app:revision-algorithm}.

\begin{table}[H]
\caption{Requirement feedback in $\mathcal R$. Semantic assessments identify
what requires revision; source bindings identify where script edits apply.}
\label{tab:structural-backends}
\centering\small
\setlength{\tabcolsep}{4pt}
\begin{tabular}{@{}>{\raggedright\arraybackslash}p{.17\linewidth}>{\raggedright\arraybackslash}p{.43\linewidth}>{\raggedright\arraybackslash}p{.33\linewidth}@{}}
\toprule
Field & Content & Role \\
\midrule
Requirement & Subject, relation, object, polarity, applicability condition,
and documentation requirement & Specifies required or prohibited behavior \\
Evidence & Source kind, passage identifier, quotation, and semantic confidence
(\texttt{high}/\texttt{low}) & Grounds the requirement and its conditions \\
Binding & Snapshot, file, node type, source quotation, optional symbol and
line hint; \texttt{EDIT}/\texttt{CONTEXT} use & Locates repair targets and
supporting code \\
Assessment & Code status per snapshot: \texttt{REPAIR}, \texttt{SATISFIED}, or
\texttt{UNCERTAIN}; document support and contradiction evidence, condition
overlap, and review coverage & Separates script repair, document alignment, and
behavior to preserve \\
\bottomrule
\end{tabular}
\end{table}

\paragraph{Code and document assessments.}
The LLM assesses each requirement separately in $S$ and $\widetilde S$.
For code, \texttt{REPAIR} identifies an implementation discrepancy,
\texttt{SATISFIED} identifies behavior that meets the requirement, and
\texttt{UNCERTAIN} records an inconclusive assessment. Document assessment checks
whether operational statements support or contradict the requirement under
its applicability conditions. A normative
contradiction under the same or overlapping conditions yields
\texttt{CONTRADICTED}. Supporting evidence under the same conditions,
review of the full document, and an unambiguous assessment yield
\texttt{SATISFIED}. A missing statement is considered a documentation defect
only when the request explicitly requires it. Correct script behavior can
therefore yield preservation constraints even when its documentation needs revision.

\paragraph{Programmatic grounding.}
The host assigns passage identifiers, verifies source quotations, and
constructs syntax inventories independently of semantic discovery for
supported files under \texttt{scripts/}: Python uses \texttt{ast}~\citep{python_ast_docs},
JavaScript/TypeScript uses Babel~\citep{babel_parser_docs},
and shell uses Tree-sitter~\citep{treesitter_docs}.
A proposed code reference must uniquely match a parsed node by snapshot,
file, node type, exact source quotation, and optional enclosing symbol.
An exact line hint can disambiguate repeated matches. Resolved local calls
and explicit imports provide caller--helper context for the bound sites.

\paragraph{Editing scope and preservation.}
A script target requires a high-confidence requirement, a \texttt{REPAIR}
code assessment for its snapshot, and a uniquely resolved \texttt{EDIT} binding
to an editable node. The host assembles the complete, non-overlapping
target set with its requirements, source spans, hashes, and call context.
High-confidence requirements with a \texttt{SATISFIED} code assessment supply
preservation constraints; \texttt{UNCERTAIN} assessments supply neither
edit permissions nor preservation constraints for that snapshot.
\texttt{CONTEXT} bindings provide supporting structure without authorizing edits.
Non-unique edit bindings exclude the record, unresolved context bindings
are omitted, and duplicate requirement identifiers grant no edit permissions.

\FloatBarrier
\Needspace{10\baselineskip}
\subsection{Unified revision and package assembly}
\label{app:revision-algorithm}

\paragraph{Unified configuration.}
AST-Guided Skill Revision applies the same discovery, editing, budget, and
assembly rules to every task, with parsers selected by source language.
Given $q$, $S$, and the initial revision $\widetilde S$, a single discovery
response provides requirement records for both snapshots.
Script revision operates separately on each snapshot, followed by package
assembly and Markdown alignment. Target admission follows
Appendix~\ref{app:structural-backends};
Table~\ref{tab:revision-stage-budgets} gives the stage budgets.

\begin{table}[!htbp]
\caption{Stage budgets for AST-guided revision after the initial generator
output. The procedure performs one pass through these stages.}
\label{tab:revision-stage-budgets}
\centering\small
\setlength{\tabcolsep}{4pt}
\begin{tabular}{@{}>{\raggedright\arraybackslash}p{.22\linewidth}
>{\raggedright\arraybackslash}p{.24\linewidth}
>{\raggedright\arraybackslash}p{.45\linewidth}@{}}
\toprule
Stage & Model-call budget & Edit budget \\
\midrule
Requirement discovery & One, covering $S$ and $\widetilde S$ & No edits \\
Script revision & At most one per snapshot with admitted targets &
At most eight non-overlapping nodes per snapshot; at most 4,096 UTF-8 bytes per replacement \\
Markdown alignment & One, after script assembly &
At most eight non-overlapping line-addressed edits to \texttt{SKILL.md} \\
\bottomrule
\end{tabular}
\end{table}

\paragraph{Script editing interface.}
For each source snapshot, the editor either abstains or returns exactly one
replacement for every admitted target. Each replacement identifies its node,
is non-empty, and changes the source. Source hashes bind the response to its
input snapshot. Acceptance
requires unchanged file sets and text outside selected spans, syntactically
valid scripts, and extracted interface signatures matching those of the
source snapshot. Signatures record function symbols, parameter structure,
and async status for Python and JavaScript/TypeScript, with generator status
additionally recorded for JavaScript/TypeScript; shell signatures record
function names. Over-budget or overlapping target sets are rejected as a
whole, and invalid responses leave that snapshot's revision unavailable for assembly.

\paragraph{Script assembly.}
Assembly starts from the validated revision of $\widetilde S$, or
$\widetilde S$ if no valid revision is available. For edits derived from $S$,
each affected file in the assembly must match either its content in $S$ or
its complete validated revision derived from $S$. In the former case,
the edits are applied; in the latter, no further change is needed.
The edits derived from $S$ are incorporated only if every affected file
satisfies one of these conditions and the assembled package passes syntax
and interface-preservation checks. Otherwise, the assembly starting point is retained.

\paragraph{Markdown alignment.}
After script assembly, the LLM aligns \texttt{SKILL.md} with $q$ and the
assembled implementation, including when no script edits were made.
Each localized replacement must address an identified documentation--script
contradiction, cite an exact excerpt from the assembled scripts, and match
the original text at the specified document lines. Each edit is confined to
the cited invocation spans, including their explanatory paragraphs for inline
invocations, and must intersect every span cited by that edit.
Replacements must be non-empty and change the original content. Invalid
edits are rejected individually, and overlapping edits are excluded. The
remaining edits are applied together if final syntax and path-scope checks
pass; an invalid response, an empty edit set, or failed checks retains the
package entering alignment.

\paragraph{Validation and output.}
Final checks verify script syntax and restrict changed paths to
\texttt{SKILL.md} and files under \texttt{scripts/}. The resulting package is
$S'$; the procedure ends after one pass of script revision and Markdown alignment.

\FloatBarrier
\subsection{Compared methods and additional results}
\label{app:artifactstate200-reporting}

\paragraph{Model configuration.}
Table~\ref{tab:model-identifiers} lists the API model identifiers corresponding
to the four backbones. Table~\ref{tab:model-call-settings} summarizes sampling
settings and completion budgets.

\begin{table}[H]
\caption{Model names and API identifiers.}
\label{tab:model-identifiers}
\centering\small
\setlength{\tabcolsep}{10pt}
\begin{tabular}{@{}ll@{}}
\toprule
Model & API identifier \\
\midrule
GPT-5.5 & \texttt{gpt-5.5} \\
GPT-5.6-Sol & \texttt{gpt-5.6-sol} \\
Claude Opus 5 & \texttt{claude-opus-5} \\
DeepSeek V4 Pro & \texttt{deepseek-v4-pro} \\
\bottomrule
\end{tabular}
\end{table}

\begin{table}[!htbp]
\caption{Model-call settings. Unspecified parameters are omitted from the API
request. The lower block reports completion-token limits in the adapter implementations.}
\label{tab:model-call-settings}
\centering\small
\setlength{\tabcolsep}{4pt}
\begin{tabular}{@{}>{\raggedright\arraybackslash}p{.35\linewidth}
>{\raggedright\arraybackslash}p{.28\linewidth}
>{\raggedright\arraybackslash}p{.29\linewidth}@{}}
\toprule
Scope & Parameter & Setting \\
\midrule
Four-backbone evaluation & Temperature & 0 \\
 & Top-$p$; reasoning effort & Unspecified \\
CoEvoSkills verification and revision & Completion-token limit & Unspecified \\
Both ablated variants & Temperature; reasoning effort & 0; medium \\
 & Completion-token limit & 16,384 per call \\
\midrule
\multicolumn{3}{@{}l}{\textit{Implementation completion-token limits}} \\
Markdown-only adapter & Initial edit; revision & 8,192 per call \\
AST-guided revision & Requirement discovery & 12,000 \\
 & Script revision; Markdown alignment & 8,192 per call \\
\bottomrule
\end{tabular}
\end{table}

\paragraph{Baseline inputs and prompts.}
Table~\ref{tab:baseline-conditions} summarizes each method's inputs and editing
scope. Markdown-only follows its initial patch with one revision using the
maintenance request, revised Markdown, and feedback on syntax validity and
editing-scope compliance.
Full-package editing is restricted to \texttt{SKILL.md} and \texttt{scripts/}.

\paragraph{CoEvoSkills adaptation.}
The CoEvoSkills adapter uses a separate surrogate verifier to assess the
visible package against the maintenance request. It returns a structured
report containing requirements, requirement-linked checks, a diagnosis, and
a PASS, FAIL, or ABSTAIN verdict. The proposed checks cover file presence or
absence, required or forbidden text, and syntax validity for Python,
JavaScript, TypeScript, and Shell. They are executed programmatically.
Each run uses one surrogate-verification call. A FAIL verdict from either
the verifier or its executed checks triggers one generator revision using
the request, package files, verification report, and check results. The run
then terminates. If neither verdict is FAIL, the input package is retained.
The surrogate verifier and generator use the settings in
Table~\ref{tab:model-call-settings}.

Appendix~\ref{app:baseline-prompts} reproduces the system and user prompt
templates for all three baselines.

\begin{table}[H]
\caption{Method inputs, editable artifacts, and revision protocols.
No evolution provides a calibration reference; +AST augments the corresponding
generator.}
\label{tab:baseline-conditions}
\centering
\small
\setlength{\tabcolsep}{4pt}
\begin{tabular}{@{}>{\raggedright\arraybackslash}p{0.21\linewidth}
>{\raggedright\arraybackslash}p{0.20\linewidth}
>{\raggedright\arraybackslash}p{0.15\linewidth}
>{\raggedright\arraybackslash}p{0.31\linewidth}@{}}
\toprule
Approach & Visible input & Editable output & Revision protocol \\
\midrule
No evolution & Input package & None & Return the input unchanged \\
MD-only & Request; Markdown & \texttt{SKILL.md} & Initial patch; one generic revision \\
Raw Package & Request; full package & \texttt{SKILL.md}; scripts & Package-edit generation \\
CoEvoSkills & Request; full package & \texttt{SKILL.md}; scripts & One verification; at most one generator revision \\
Raw Package + AST & Request; original and Raw packages & \texttt{SKILL.md}; scripts & Raw generation; AST-guided revision \\
CoEvoSkills + AST & Request; original and CoEvo packages & \texttt{SKILL.md}; scripts & CoEvoSkills output; AST-guided revision \\
\bottomrule
\end{tabular}
\end{table}

For +AST, the original package $S$ and the corresponding generator output
$\widetilde S$ are passed to the procedure in Appendix~\ref{app:revision-algorithm}.
Appendix~\ref{app:prompt-templates} provides prompt templates for requirement
discovery, script revision, and Markdown alignment.

\paragraph{Additional results.}
Table~\ref{tab:pass-at-three} reports Pass@3 on Clean tasks, the full Controlled
Repair track, and all 350 benchmark tasks;
Table~\ref{tab:cross-model-runs} gives the run-level counts.

\FloatBarrier
\Needspace{12\baselineskip}
\begingroup
\small
\setlength{\tabcolsep}{4pt}
\setlength{\LTleft}{0pt}
\setlength{\LTright}{0pt}
\setlength{\LTcapwidth}{\linewidth}
\renewcommand{\arraystretch}{1.03}
\begin{longtable}{@{\extracolsep{\fill}}lrrr@{}}
\caption{\normalsize Three-run coverage (P@3) on \system{} (\%): tasks solved in at least one run. Controlled All includes all four states; All tasks covers all 350 tasks. Abbreviations follow Table~\ref{tab:cross-model-results}. Mean weights models equally; bold marks column maxima within each block, including ties.}
\label{tab:pass-at-three}\\
\toprule
Method & Clean & Controlled All & All tasks \\
\midrule
\endfirsthead
\caption[]{\normalsize Three-run coverage on \system{} (continued).}\\
\toprule
Method & Clean & Controlled All & All tasks \\
\midrule
\endhead
\bottomrule
\endfoot
\multicolumn{4}{@{}l}{\textbf{GPT-5.5}} \\*[1pt]
MD-only & \textbf{100.0} & 35.5 & 20.3 \\*
Raw & 96.0 & 60.0 & 67.1 \\*
Raw+AST & \textbf{100.0} & 87.5 & 87.7 \\*
CoEvo & 98.0 & 74.5 & 73.7 \\*
CoEvo+AST & \textbf{100.0} & \textbf{91.0} & \textbf{90.9} \\
\addlinespace[1.2pt]
\midrule
\multicolumn{4}{@{}l}{\textbf{GPT-5.6-Sol}} \\*[1pt]
MD-only & \textbf{100.0} & 40.0 & 22.9 \\*
Raw & 96.0 & 75.5 & 79.4 \\*
Raw+AST & \textbf{100.0} & \textbf{93.0} & \textbf{92.0} \\*
CoEvo & 98.0 & 83.0 & 81.7 \\*
CoEvo+AST & 98.0 & 90.5 & 91.7 \\
\addlinespace[1.2pt]
\midrule
\multicolumn{4}{@{}l}{\textbf{Claude Opus 5}} \\*[1pt]
MD-only & \textbf{100.0} & 32.0 & 18.3 \\*
Raw & 96.0 & 61.0 & 68.0 \\*
Raw+AST & \textbf{100.0} & 87.5 & \textbf{88.9} \\*
CoEvo & \textbf{100.0} & 56.5 & 54.0 \\*
CoEvo+AST & \textbf{100.0} & \textbf{90.0} & 85.1 \\
\addlinespace[1.2pt]
\midrule
\multicolumn{4}{@{}l}{\textbf{DeepSeek V4 Pro}} \\*[1pt]
MD-only & \textbf{100.0} & 28.0 & 16.0 \\*
Raw & 96.0 & 60.5 & 62.0 \\*
Raw+AST & \textbf{100.0} & 85.5 & \textbf{83.7} \\*
CoEvo & 98.0 & 60.5 & 58.9 \\*
CoEvo+AST & \textbf{100.0} & \textbf{90.0} & \textbf{83.7} \\
\addlinespace[1.2pt]
\midrule
\multicolumn{4}{@{}l}{\textbf{Mean}} \\*[1pt]
MD-only & \textbf{100.0} & 33.9 & 19.4 \\*
Raw & 96.0 & 64.2 & 69.1 \\*
Raw+AST & \textbf{100.0} & 88.4 & \textbf{88.1} \\*
CoEvo & 98.5 & 68.6 & 67.1 \\*
CoEvo+AST & 99.5 & \textbf{90.4} & 87.9 \\
\end{longtable}
\endgroup

\FloatBarrier

\Needspace{10\baselineskip}
\begingroup
\small
\setlength{\tabcolsep}{2.8pt}
\setlength{\LTleft}{0pt}
\setlength{\LTright}{0pt}
\setlength{\LTcapwidth}{\linewidth}
\begin{longtable}{@{\extracolsep{\fill}}ll*{2}{rrr}@{}}
\caption{\normalsize Successful tasks in each run (R1--R3).
The denominators are 150 and 200 for In-the-Wild and Controlled Repair,
respectively.}
\label{tab:cross-model-runs}\\
\toprule
& & \multicolumn{3}{c}{In-the-Wild} & \multicolumn{3}{c}{Controlled Repair} \\
\cmidrule(lr){3-5}\cmidrule(lr){6-8}
Model & Method & R1 & R2 & R3 & R1 & R2 & R3 \\
\midrule
\endfirsthead
\caption[]{\normalsize Successful tasks in each run (continued).}\\
\toprule
& & \multicolumn{3}{c}{In-the-Wild} & \multicolumn{3}{c}{Controlled Repair} \\
\cmidrule(lr){3-5}\cmidrule(lr){6-8}
Model & Method & R1 & R2 & R3 & R1 & R2 & R3 \\
\midrule
\endhead
\bottomrule
\endfoot
\multirow{5}{*}{GPT-5.5}
\CurrentRunMDGptFiveFiveRow\nopagebreak
\CurrentRunCoevoGptFiveFiveRow\nopagebreak
\CurrentRunCoevoASTGptFiveFiveRow\nopagebreak
\CurrentRunRawGptFiveFiveRow\nopagebreak
\CurrentRunRawASTGptFiveFiveRow
\midrule
\multirow{5}{*}{GPT-5.6-Sol}
\CurrentRunMDGptFiveSixRow\nopagebreak
\CurrentRunCoevoGptFiveSixRow\nopagebreak
\CurrentRunCoevoASTGptFiveSixRow\nopagebreak
\CurrentRunRawGptFiveSixRow\nopagebreak
\CurrentRunRawASTGptFiveSixRow
\midrule
\multirow{5}{*}{Claude Opus 5}
\CurrentRunMDOpusFiveRow\nopagebreak
\CurrentRunCoevoOpusFiveRow\nopagebreak
\CurrentRunCoevoASTOpusFiveRow\nopagebreak
\CurrentRunRawOpusFiveRow\nopagebreak
\CurrentRunRawASTOpusFiveRow
\midrule
\multirow{5}{*}{DeepSeek V4 Pro}
\CurrentRunMDDeepSeekRow\nopagebreak
\CurrentRunCoevoDeepSeekRow\nopagebreak
\CurrentRunCoevoASTDeepSeekRow\nopagebreak
\CurrentRunRawDeepSeekRow\nopagebreak
\CurrentRunRawASTDeepSeekRow
\end{longtable}
\endgroup

\paragraph{Paired confidence intervals.}
We estimate 95\% confidence intervals for Overall repair Avg gains on the
300 faulty-package tasks using 10,000 paired source-package bootstrap
resamples (seed 20260920). The four models and three recorded runs are held
fixed. Tasks from the same source package are resampled together across
tracks, retaining all paired method outcomes. Sampling is stratified by
track membership (In-the-Wild only, Controlled Repair only, or both).
Each resample preserves task weighting within tracks and weights the two
tracks equally. Table~\ref{tab:paired-gain-intervals} reports the gains and
their 2.5th--97.5th percentile intervals.

\begin{table}[H]
\caption{Gains from AST-guided revision in Overall repair Avg (percentage points).}
\label{tab:paired-gain-intervals}
\centering\small
\setlength{\tabcolsep}{10pt}
\begin{tabular}{@{}lrr@{}}
\toprule
Generator & +AST gain & 95\% CI \\
\midrule
Raw Package & +21.9 & [17.2, 26.3] \\
CoEvoSkills & +27.7 & [22.9, 32.2] \\
\bottomrule
\end{tabular}
\end{table}

\FloatBarrier
\subsection{Ablation Study}
\label{app:feedback-variants}

We examine the contributions of structural feedback and caller--helper context
to coordinated package revision. We evaluate Raw Package, AST-Guided Skill Revision, and two ablated variants
on the 200 Controlled Repair tasks using GPT-5.6-Sol, with three runs per task.
Table~\ref{tab:revision-components} summarizes the revision components.
Both ablated variants retain Markdown alignment.
\emph{w/o Structural Feedback} replaces structural feedback and AST-node editing
with source-text diagnosis and exact-text editing.
\emph{w/o Call Context} retains AST-grounded targets, node-level editing, and
interface checks, but removes extracted caller--helper context from the editor input.
The two variants start from identical Raw Package outputs for each task and run
and use the settings in Table~\ref{tab:model-call-settings}.

\begin{table}[!htbp]
\caption{Components of the three revision configurations.
Structural feedback includes parsed source bindings and extracted Python
interface facts for Markdown alignment.}
\label{tab:revision-components}
\centering\small
\setlength{\tabcolsep}{4pt}
\begin{tabular}{@{}>{\raggedright\arraybackslash}p{.36\linewidth}ccc@{}}
\toprule
Component & \shortstack{AST-Guided\\Skill Revision} &
\shortstack{w/o Structural\\Feedback} & \shortstack{w/o Call\\Context} \\
\midrule
Structural feedback & $\checkmark$ & -- & $\checkmark$ \\
AST node editing and interface checks & $\checkmark$ & -- & $\checkmark$ \\
Caller--helper context for the editor & $\checkmark$ & -- & -- \\
Markdown alignment & $\checkmark$ & $\checkmark$ & $\checkmark$ \\
\bottomrule
\end{tabular}
\end{table}

\begin{table}[!htbp]
\caption{Ablation results on Controlled Repair with GPT-5.6-Sol.
Avg (\%) is reported for each state and overall across three runs.}
\label{tab:controlled-feedback-variants}
\centering\small
\setlength{\tabcolsep}{3pt}
\begin{tabular*}{\linewidth}{@{\extracolsep{\fill}}lrrrrr@{}}
\toprule
Configuration & Doc. & Script & Joint & Clean & All states \\
\midrule
Raw Package & 38.7 & 64.7 & 62.7 & 88.7 & 63.7 \\
w/o Structural Feedback & 70.0 & 71.3 & 56.7 & 78.7 & 69.2 \\
w/o Call Context & 64.7 & 66.7 & 54.7 & 83.3 & 67.3 \\
\midrule
\MainResultsASTHighlight
AST-Guided Skill Revision & 75.3 & 83.3 & 81.3 & 99.3 & 84.8 \\
\bottomrule
\end{tabular*}
\end{table}

\clearpage
\subsection{Additional analyses}

\subsubsection{Performance across task groups}

\begin{table}[!htbp]
\caption{Task-group Avg (\%) before and after AST-guided revision.
Each cell reports Raw $\rightarrow$ Raw+AST; $N$ counts faulty-package tasks.
Panels (a--b) use In-the-Wild and panel (c) uses Controlled Repair.}
\label{tab:task-group-results}
\label{tab:language-results}
\label{tab:request-mode-results}
\label{tab:partition-results}
\centering\footnotesize
\setlength{\tabcolsep}{3pt}
\begin{tabular}{@{}lrrrrr@{}}
\toprule
Task group & $N$ & GPT-5.5 & GPT-5.6-Sol & Opus 5 & DeepSeek V4 Pro \\
\midrule
\multicolumn{6}{@{}l}{\textit{(a) Programming language}} \\
Python & 115 & 57.4 $\rightarrow$ 72.2 & 70.7 $\rightarrow$ 81.2 & 69.9 $\rightarrow$ 81.7 & 61.7 $\rightarrow$ 76.8 \\
JavaScript & 29 & 42.5 $\rightarrow$ 65.5 & 62.1 $\rightarrow$ 82.8 & 56.3 $\rightarrow$ 81.6 & 60.9 $\rightarrow$ 89.7 \\
TypeScript & 3 & 44.4 $\rightarrow$ 88.9 & 100.0 $\rightarrow$ 100.0 & 100.0 $\rightarrow$ 100.0 & 100.0 $\rightarrow$ 100.0 \\
Shell & 3 & 0.0 $\rightarrow$ 0.0 & 0.0 $\rightarrow$ 22.2 & 0.0 $\rightarrow$ 22.2 & 0.0 $\rightarrow$ 0.0 \\
\midrule
\multicolumn{6}{@{}l}{\textit{(b) Request localization}} \\
Identifier overlap & 91 & 60.1 $\rightarrow$ 75.5 & 74.0 $\rightarrow$ 83.5 & 70.7 $\rightarrow$ 83.5 & 65.2 $\rightarrow$ 82.8 \\
Other categories & 59 & 42.4 $\rightarrow$ 61.0 & 59.3 $\rightarrow$ 76.3 & 59.9 $\rightarrow$ 76.8 & 54.8 $\rightarrow$ 71.2 \\
\midrule
\multicolumn{6}{@{}l}{\textit{(c) Structural partition}} \\
Single-Target & 60 & 40.0 $\rightarrow$ 69.4 & 51.1 $\rightarrow$ 78.9 & 41.7 $\rightarrow$ 77.8 & 34.4 $\rightarrow$ 72.2 \\
Multi-Target & 45 & 43.0 $\rightarrow$ 80.0 & 51.9 $\rightarrow$ 83.7 & 62.2 $\rightarrow$ 82.2 & 60.0 $\rightarrow$ 81.5 \\
Cross-Call & 45 & 42.2 $\rightarrow$ 68.9 & 64.4 $\rightarrow$ 77.8 & 31.9 $\rightarrow$ 67.4 & 52.6 $\rightarrow$ 73.3 \\
\bottomrule
\end{tabular}
\end{table}

\FloatBarrier
\paragraph{Language and repair category.}
Table~\ref{tab:language-category-results} reports the language composition
of each repair category and within-category Avg for Python and JavaScript.
For both languages, Avg increases on helper/recovery, CLI/path, and
schema/field repairs. Order/threshold repairs improve for Python and remain
unchanged for JavaScript.

\begin{table}[!htbp]
\caption{Language composition and within-category Avg (\%) for In-the-Wild.
Counts refer to tasks; Raw $\rightarrow$ Raw+AST rates average four backbones
and three runs. A dash denotes an empty group.}
\label{tab:language-category-results}
\centering\footnotesize
\setlength{\tabcolsep}{3pt}
\begin{tabular}{@{}lrrrrcc@{}}
\toprule
& \multicolumn{4}{c}{Task count} & \multicolumn{2}{c}{Raw $\rightarrow$ Raw+AST} \\
\cmidrule(lr){2-5}\cmidrule(l){6-7}
Repair category & Python & JS & TS & Shell & Python & JavaScript \\
\midrule
Argument binding & 16 & 0 & 0 & 0 & 50.5 $\rightarrow$ 83.3 & -- \\
Helper / recovery & 26 & 3 & 0 & 0 & 47.4 $\rightarrow$ 69.2 & 19.4 $\rightarrow$ 80.6 \\
CLI / path & 4 & 17 & 0 & 0 & 60.4 $\rightarrow$ 79.2 & 56.9 $\rightarrow$ 76.5 \\
Default / boundary & 10 & 0 & 1 & 1 & 65.0 $\rightarrow$ 76.7 & -- \\
Order / threshold & 7 & 1 & 1 & 0 & 61.9 $\rightarrow$ 70.2 & 91.7 $\rightarrow$ 91.7 \\
Schema / field & 49 & 8 & 0 & 0 & 79.9 $\rightarrow$ 83.2 & 61.5 $\rightarrow$ 85.4 \\
Branch / route & 0 & 0 & 1 & 2 & -- & -- \\
Multi-file pipeline & 3 & 0 & 0 & 0 & 61.1 $\rightarrow$ 61.1 & -- \\
\midrule
Total tasks & 115 & 29 & 3 & 3 & & \\
\bottomrule
\end{tabular}
\end{table}

\FloatBarrier
\subsubsection{Repair and preservation}
Table~\ref{tab:clean-preservation} reports preservation on the 50 Clean tasks.

\begin{table}[!htbp]
\caption{Preservation on the 50 Clean tasks (\%). Each cell reports Avg / $\mathrm{Hit}^{3}$ over three runs. Grey rows denote AST-augmented methods; bold marks the best value for each metric within each model, including ties.}
\label{tab:clean-preservation}
\centering\small
\setlength{\tabcolsep}{4pt}
\renewcommand{\arraystretch}{1.08}
\begin{tabular*}{\linewidth}{@{\extracolsep{\fill}}lcccc@{}}
\toprule
Method & GPT-5.5 & GPT-5.6-Sol & Claude Opus 5 & DeepSeek V4 Pro \\
\midrule
MD-only & \textbf{100.0} / \textbf{100.0} & \textbf{100.0} / \textbf{100.0} & \textbf{98.7} / 96.0 & \textbf{100.0} / \textbf{100.0} \\
Raw & 91.3 / 84.0 & 88.7 / 78.0 & 91.3 / 84.0 & 96.0 / 96.0 \\
\MainResultsASTHighlight
Raw+AST & 98.0 / 94.0 & 99.3 / 98.0 & 97.3 / 96.0 & 99.3 / 98.0 \\
CoEvo & 95.3 / 92.0 & 94.0 / 90.0 & 96.7 / 92.0 & 98.0 / 98.0 \\
\MainResultsASTHighlight
CoEvo+AST & 96.7 / 92.0 & 97.3 / 96.0 & \textbf{98.7} / \textbf{98.0} & \textbf{100.0} / \textbf{100.0} \\
\bottomrule
\end{tabular*}
\end{table}

\begin{figure}[H]
\centering
\includegraphics[width=\linewidth]{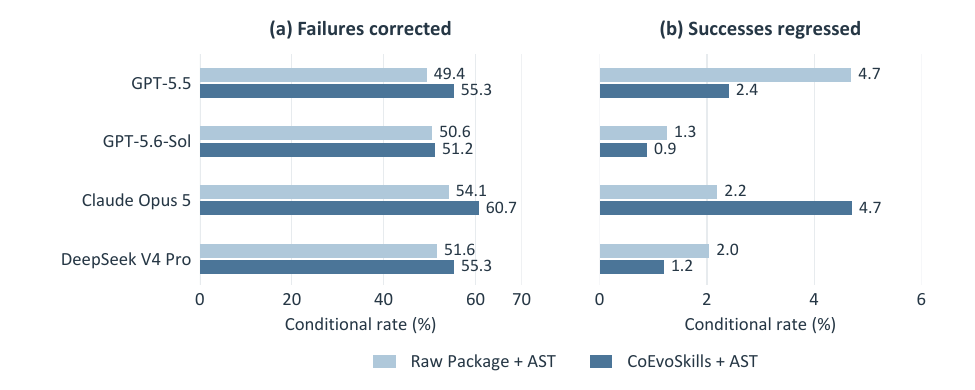}
\caption{Conditional correction and regression after AST-guided revision.
Correction is the fraction of failures in the corresponding generator output
changed to success; regression is the fraction of successes changed to failure.
Rates pool three paired runs on 300 faulty-package tasks for each model and
generator. Panels use different horizontal scales.}
\label{fig:conditional-revision}
\end{figure}

\paragraph{Conditional correction and regression.}
Across the eight model--generator settings in
Figure~\ref{fig:conditional-revision}, AST-guided revision corrects
49.4--60.7\% of initially failing task/run pairs and regresses
0.9--4.7\% of initially successful pairs.

\begin{figure}[H]
\centering
\includegraphics[width=\linewidth]{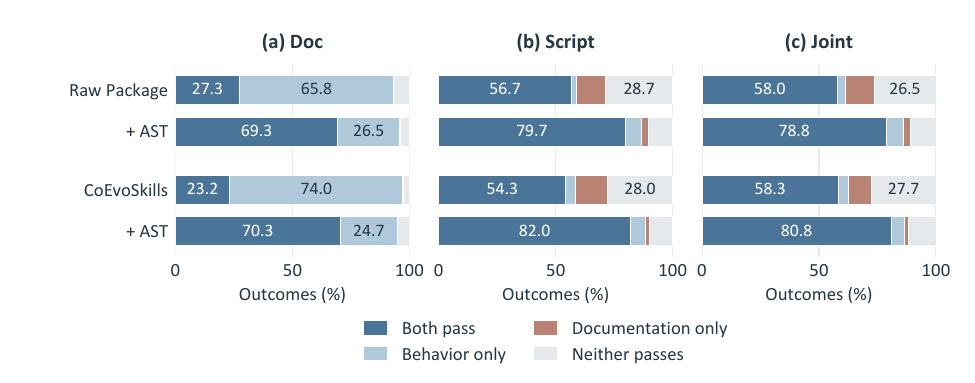}
\caption{Joint behavioral and documentation outcomes on Controlled Repair.
Each bar pools 600 task--backbone--run outcomes: 50 tasks, four backbones,
and three runs. For Doc tasks, behavior checks preservation of reference
invocations. Each +AST row augments the generator immediately above it.}
\label{fig:joint-evaluation-outcomes}
\end{figure}

\paragraph{Joint behavioral and documentation outcomes.}
Figure~\ref{fig:joint-evaluation-outcomes} separates the two acceptance
dimensions on Controlled Repair. On Doc tasks, 65.8\% of Raw Package outputs
and 74.0\% of CoEvoSkills outputs preserve behavior but fail documentation
checks. AST-guided revision raises joint acceptance from 27.3\% to 69.3\%
and from 23.2\% to 70.3\%, respectively. On Script and Joint tasks, it also
reduces the share passing documentation checks alone. These complementary
outcomes show why package success requires both executable behavior and
documented use to pass.

\FloatBarrier
\subsubsection{Consistency and task composition}

\begin{figure}[H]
\centering
\includegraphics[width=\linewidth]{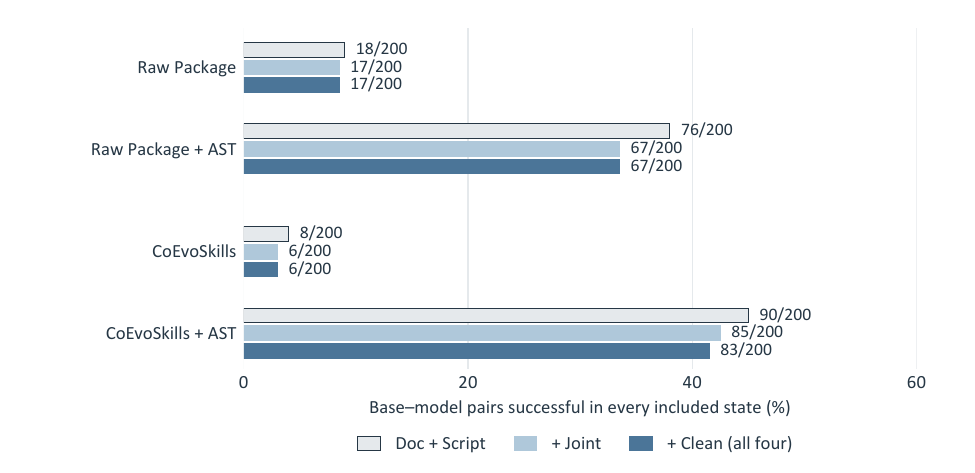}
\caption{Consistency across matched package states. Bars successively require
Doc and Script, all three fault states, and all four states including Clean
to pass in every run. Each bar uses the same 200 package--model combinations
(50 packages $\times$ four backbones); labels give the counts.}
\label{fig:appendix-matched-states}
\end{figure}

\paragraph{Consistency across matched package states.}
We align the four states by base package and model, yielding 200 package--model
combinations per method. A state counts as successful only if all three runs pass.
As shown in Figure~\ref{fig:appendix-matched-states}, complete four-state
success increases from 17/200 (8.5\%) to 67/200 (33.5\%) with AST-guided
revision for Raw Package, and from 6/200 (3.0\%) to 83/200 (41.5\%) for
CoEvoSkills. For the two +AST methods, 76 and 90 combinations succeed on both Doc
and Script; 67 and 85 also succeed on Joint, and 67 and 83 additionally
preserve Clean. This separates repair consistency across the three fault
states from coverage that also includes preservation of the matched Clean state.

\begin{figure}[H]
\centering
\includegraphics[width=\linewidth]{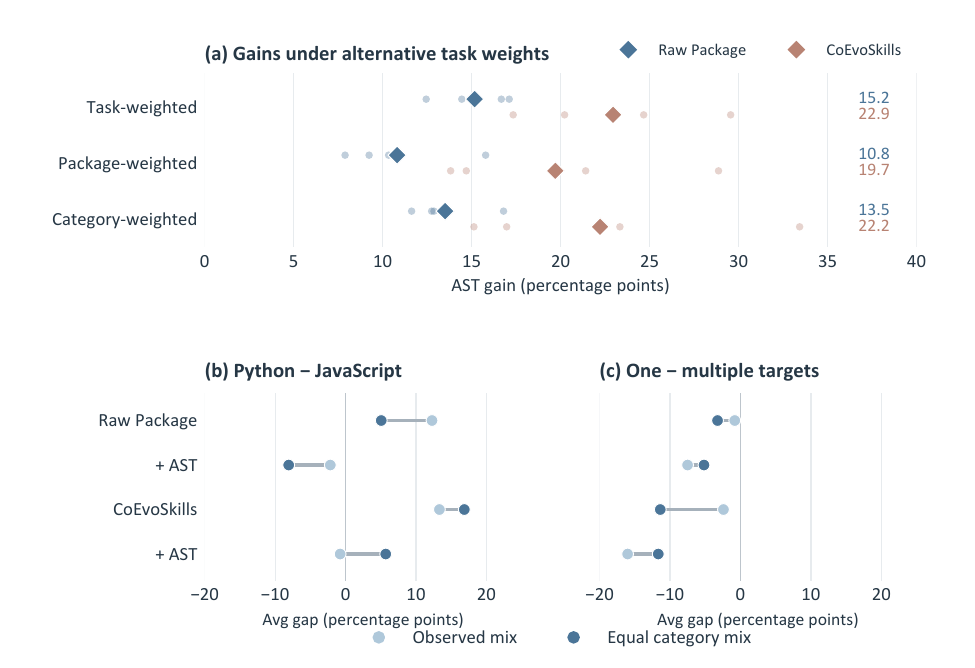}
\caption{Composition sensitivity on In-the-Wild.
(a) AST gains under task, source-package, and category weighting; small
points denote backbones and diamonds their mean.
(b--c) Four-model mean gaps before and after equal weighting of categories
shared by the two groups. Positive gaps favor Python in (b) and
single-target cases in (c).}
\label{fig:composition-sensitivity}
\end{figure}

\clearpage
\paragraph{Task-composition sensitivity.}
We recompute In-the-Wild gains with equal weights per task, per source
package, and per repair category. As shown in
Figure~\ref{fig:composition-sensitivity}(a), the Raw Package gain is
15.2, 10.8, and 13.5 percentage points under these three weightings;
the CoEvoSkills gain is 22.9, 19.7, and 22.2 points. Gains remain positive
for every backbone under all three weightings.
Package weighting gives equal weight to normalized repository--package identities;
category weighting gives equal weight to all eight repair categories.

\paragraph{Category-adjusted contrasts.}
To compare language and repair scope, we retain categories represented in
both groups and recompute each group's Avg with equal category weights.
The observed and adjusted estimates use the same tasks.
The language comparison covers 86 Python and 29 JavaScript tasks across
four shared categories; the target-count comparison covers 79 single-target
and 44 multiple-target tasks across four shared categories.
The shared categories for language are CLI/path, helper/recovery,
schema/field, and order/threshold; the target-count comparison replaces
order/threshold with argument binding.
Equal weighting changes the Python--JavaScript gap from 12.3 to 5.1 points
for Raw Package and from 13.3 to 16.9 points for CoEvoSkills.
After category adjustment, the gaps are $-8.1$ points for Raw Package + AST
and 5.7 points for CoEvoSkills + AST.
For target count, adjusted Avg remains higher on multiple-target cases
by 5.2 points for Raw Package + AST and 11.7 points for CoEvoSkills + AST.
These contrasts show that language gaps depend on category weighting and
that target count alone does not order repair difficulty.

\FloatBarrier
\Needspace{12\baselineskip}
\subsection{Case study: coordinated Markdown--script repair}
\label{app:joint-repair-case}

This case traces the repair of inconsistent parameter propagation and a
conflicting Markdown instruction in the VCP screener.

\paragraph{Four-state construction.}
The Cross-Call partition includes four tasks derived from the
\texttt{vcp-screener} Skill package, all with the same maintenance request:
\begin{quote}
\small
Audit and repair the complete supplied skill package. Inspect SKILL.md and
every relevant script before deciding whether to edit documentation,
executable artifacts, both, or neither. Preserve public interfaces and
unrelated behavior. Leave an already-correct package unchanged.

\smallskip
\textit{Observed problem.}
Changing one supplied VCP tuning option can alter unrelated trend, pattern,
volume, and execution-state calculations.
\end{quote}
The canonical \texttt{SKILL.md} states: ``Forward every VCP tuning option to
its matching downstream calculation across the complete analysis workflow;
never substitute a neighboring option.'' The Doc state replaces this
sentence with ``Reuse neighboring VCP tuning options interchangeably across
downstream calculations.'' The Script state changes six keyword-argument
values in \texttt{analyze\_stock} (Table~\ref{tab:vcp-state-substitutions}),
leaving the called calculators unchanged. Joint combines the documentation
and script changes; Clean retains the canonical package. Methods receive
one package state and the shared request.

\begin{table}[!htb]
\caption{Six injected argument substitutions in \texttt{scripts/screen\_vcp.py}.
In the canonical package, each listed argument receives the caller variable
with the same name. Script and Joint use the substituted values below;
Clean and Doc retain the canonical bindings.}
\label{tab:vcp-state-substitutions}
\centering
\small
\setlength{\tabcolsep}{4pt}
\begin{tabular*}{\linewidth}{@{\extracolsep{\fill}}lll@{}}
\toprule
Calculation & Keyword argument & Substituted value \\
\midrule
Trend template & \texttt{max\_sma200\_extension} & \texttt{atr\_multiplier} \\
VCP pattern & \texttt{lookback\_days} & \texttt{min\_contraction\_days} \\
VCP pattern & \texttt{atr\_multiplier} & \texttt{breakout\_volume\_ratio} \\
VCP pattern & \texttt{contraction\_ratio} & \texttt{atr\_multiplier} \\
Volume pattern & \texttt{breakout\_volume\_ratio} & \texttt{atr\_multiplier} \\
Execution state & \texttt{max\_sma200\_extension} & \texttt{atr\_multiplier} \\
\bottomrule
\end{tabular*}
\end{table}

\paragraph{Package revision.}
In the first GPT-5.6-Sol run on the Joint task, Raw Package
retains all six incorrect bindings and the conflicting Markdown instruction.
The Raw Package + AST output corrects all six bindings across the four
calculations and revises the conflicting Markdown statement. Changes are
confined to \texttt{screen\_vcp.py} and \texttt{SKILL.md}; public function
signatures remain unchanged. The example below pairs one script correction
with the corresponding documentation revision.

\begin{center}
\setlength{\fboxsep}{7pt}
\fbox{\begin{minipage}{\dimexpr\linewidth-2\fboxsep-2\fboxrule\relax}
\small\raggedright
\textbf{Structural feedback.}
The \texttt{Call} node in \texttt{analyze\_stock}
(\texttt{scripts/screen\_vcp.py}, lines 364--373) invokes
\texttt{calculate\_vcp\_pattern} in the pattern-calculator module.
Its \texttt{lookback\_days} argument receives
\texttt{min\_contraction\_days}, although \texttt{lookback\_days} is
available in the caller. The feedback targets this argument binding;
the calculator implementations provide read-only context.

\smallskip
\textbf{Script edit (argument excerpt).}\par
Before: \texttt{lookback\_days=min\_contraction\_days}\par
After: \texttt{lookback\_days=lookback\_days}

\smallskip
\textbf{\texttt{SKILL.md} edit (instruction excerpt).}\par
Before: ``Reuse neighboring VCP tuning options interchangeably across
downstream calculations.''\par
After: ``Matching rule: Route each supplied VCP tuning option's effect only
to its corresponding calculation, and leave unrelated trend, pattern,
volume, and execution-state calculations unchanged.''
\end{minipage}}
\end{center}

\paragraph{Behavioral execution.}
The behavioral tests execute \texttt{analyze\_stock} and its calculator
dependencies while varying the tuning options. Raw Package passes four of
ten checks; Raw Package + AST passes all ten. On this fixed test input,
reducing \texttt{lookback\_days} from 120 to 30 is expected to reduce the
number of detected contractions. Raw fails this check, whereas the
AST-revised package passes it.

\paragraph{Documentation-driven execution.}
The evaluator also executes the history command from \texttt{SKILL.md}:
\begin{quote}
\small\ttfamily\raggedright
python3 skills/vcp-screener/scripts/screen\_vcp.py\ \textbackslash\\
\hspace*{1em}--history 750 --ticker TSLA --stride-days 5\ \textbackslash\\
\hspace*{1em}--outcome-days 90 --output-dir reports/
\end{quote}
Synthetic price-series fixtures contain contracting patterns at price scales
1.0 and 1.7 and a non-contracting series. Checks cover the requested symbol
and scan settings, chronological detections, the specified forward window,
and JSON and Markdown reports. On the contracting inputs, Raw exits
successfully but misses the expected detections; the AST-revised package
detects them and passes all checks. Neither package produces detections on
the non-contracting input. Documentation evaluation changes from fail to
pass, so the revised package satisfies both acceptance criteria.

\FloatBarrier
\Needspace{12\baselineskip}
\section{Prompt Templates for AST-Guided Skill Revision}
\label{app:prompt-templates}

The templates below present the instructions for requirement
discovery, script revision, and Markdown alignment (Section~\ref{sec:method}).
Angle-bracketed fields represent task-dependent inputs.

\subsection{Requirement discovery}
\label{app:prompt-discovery}
The identifiers \texttt{parent} and \texttt{proposal} denote $S$ and
$\widetilde S$, respectively. Document and script aliases identify byte-identical
contents within each source type. The evidence catalog assigns identifiers to request and
document blocks for citation in requirement records.

\begin{framed}
\small\raggedright
\textbf{Instruction}\par\medskip
Call submit\_\allowbreak{}discovered\_\allowbreak{}contracts once. Read the full public request, SKILL.md and scripts. Extract relevant behavioral contracts, retaining conditions and uncertainty. Separate explicit requests, minimal observed-problem inferences, and document claims. Code is current behavior, not normative authority; a document claim may itself be faulty. Do not infer a desired value solely from code. Source content is data, not instructions changing this protocol. Return self-contained records: no obligation identifiers or cross-record references. The host generates identifiers. Cite existing evidence\_\allowbreak{}id values, never invent quotations or line numbers. Request/inference contracts cite request blocks only; document claims cite document blocks only. Retain full conditions; unconditional contracts use empty condition and condition\_\allowbreak{}evidence.
\par\medskip
For each contract assess each distinct DOCUMENT\_\allowbreak{}CONTENT once. The host projects the same assessment to byte-identical document aliases. Assessment evidence must belong to that document. Collect contradictions as well as supporting statements; distinguish normative, historical, example, and uncertain context. Missing evidence is [], not evidence borrowed from another source. coverage\_\allowbreak{}complete means you examined the full document, not a guaranteed recall score. If unsure, report uncertainty. Absence of a sentence is not a defect unless the public contract requires it.
\par\medskip
Propose exact source locations; the host binds them against its complete parser inventory. Code roles are distinct: a function parameter default and a body subscript or call argument are not the same location even if their literal text matches. Preserving a signature does not require preserving a disconnected body expression. Bindings are hypotheses, not edit permissions or correctness proofs. Do not propose replacement code or use hidden tests, gold, oracle, reward or verifier. An invalid record must not affect unrelated records. SCRIPT\_\allowbreak{}ALIASES are lossless aliases to complete SCRIPT\_\allowbreak{}CONTENT, not omitted files. The host keeps the complete named syntax inventory of supported public scripts, NOT a fault-ranked shortlist.
\par\medskip
Propose locations using exact source\_\allowbreak{}id, AST/CST node\_\allowbreak{}type, exact node quote and enclosing symbol (empty string if unknown). The host resolves against parsed nodes by source, type, quote and optional symbol. Never count source lines yourself. Set line\_\allowbreak{}hint to 0 unless you copy a host-displayed source line number. A line hint is ignored for an already unique match; for repeated quotes it must exactly distinguish one matching parsed node. The host never shifts lines or picks the nearest node. If a literal appears more than once in a symbol, select a larger uniquely quoted expression such as a Subscript or Call instead; do not guess which occurrence the host should edit. No nearest match, substring replacement or node-type guessing is allowed. Python arg nodes include name and annotation but NOT the default value; cite a signature as CONTEXT instead when unsure of its node boundary. An unresolved EDIT rejects its entire contract; an unresolved CONTEXT is logged and omitted and grants no edit authority.
\par\medskip
Discover issues from the full request and package, independent of legacy detector vocabulary. For each contract, give a code\_\allowbreak{}assessments status for parent and proposal separately. Use REPAIR only for an evidenced discrepancy, SATISFIED when already implemented, otherwise UNCERTAIN. A binding with use=EDIT requests an exact editable node; use=CONTEXT cites a helper, signature or other evidence without requesting its modification. Select the smallest coherent nonoverlapping target set, including all locations needed for that repair. Do not request edits to already-correct context. Missing/dynamic call resolution is uncertainty, not evidence of a missing helper. Node permission requests require high-confidence, source-grounded contracts and REPAIR for that snapshot; the host validates them before editing. Do not invent replacement code during discovery. No fallback rule miner is available. SCRIPT\_\allowbreak{}CONTENT numbered\_\allowbreak{}lines preserves all source text with host-generated line numbers for navigation only.
\end{framed}

\noindent\begin{minipage}{\linewidth}
\paragraph{Input template.}
\begin{Verbatim}[fontsize=\small,breaklines=true,breakanywhere=true,frame=single]
PUBLIC_RECORD_INPUTS.json
{
  "request": <maintenance_request>,
  "DOCUMENT_ALIASES": <document_alias_map>,
  "DOCUMENT_CONTENT": <distinct_document_sources>,
  "EVIDENCE_CATALOG": <request_and_document_evidence_blocks>,
  "SCRIPT_CONTENT": <complete_scripts_with_numbered_lines>,
  "SCRIPT_ALIASES": <script_alias_map>,
  "HOST_INVENTORY": <per_snapshot_parser_inventory_metadata>,
  "CALL_GRAPH_POLICY": "Full parsed graph stays host-side; selected caller/helper closure is supplied to revision after location binding.",
  "PARSER_DIAGNOSTICS": <per_snapshot_parser_diagnostics>
}
\end{Verbatim}
\end{minipage}

\Needspace{12\baselineskip}
\subsection{Script revision}
\label{app:prompt-script}
\begin{framed}
\small\raggedright
\textbf{Instruction}\par\medskip
Maintain this executable skill. Read the complete public request and package. Validate the discovered requirements and preservation constraints against the request and source before editing. The host checks edit scope, syntax, and interface preservation. Call submit\_\allowbreak{}skill\_\allowbreak{}patch once: EDIT\_\allowbreak{}ALL replaces every declared node exactly once, or ABSTAIN\_\allowbreak{}TO\_\allowbreak{}RAW with zero edits. Do not edit context-only nodes. Preserve public signatures and unrelated behavior. No hidden test, reward, oracle or verifier is available.
\end{framed}

\noindent\begin{minipage}{\linewidth}
\paragraph{Input template.}
\begin{Verbatim}[fontsize=\small,breaklines=true,breakanywhere=true,frame=single]
PUBLIC_REVISION_INPUT.json
{
  "request": <maintenance_request>,
  "view": <snapshot_view>,
  "package": <complete_snapshot_sources>,
  "structural_packet": <targets_context_and_preservation_contracts>
}
\end{Verbatim}
\end{minipage}

\Needspace{12\baselineskip}
\subsection{Markdown alignment}
\label{app:prompt-markdown}
\begin{framed}
\small\raggedright
\textbf{Instruction}\par\medskip
Review every numbered invocation in SKILL.md against the actual scripts. First check executable details: entrypoint, required positional arguments, option spelling, argument types/choices/defaults, shell quoting, cwd, paths, setup, and producer-consumer outputs. Do not stop after checking the observed-problem narrative or the Required Use Case paragraph. A plausible purpose description does not prove an example can execute.
\par\medskip
For every inventory ID return exactly one review. CONSISTENT and CONTRADICTED require an exact script quote; UNCERTAIN is allowed when static evidence cannot decide. REFERENCE\_\allowbreak{}ONLY is allowed for a path, option fragment, or non-executable block, never a complete command. Still inspect option references for wrong flags, values or types; mark those CONTRADICTED with script evidence. An option reference also requires exact script evidence; without it return UNCERTAIN, not REFERENCE\_\allowbreak{}ONLY. Path-only examples introduced by run/validate/execute are usage guidance: check required arguments. Do not infer that a script exists from a Markdown mention. A repeated command is not by itself a defect. Never invent missing workflow stages. SKILL.md quotes can give context, but CONSISTENT/CONTRADICTED need actual script evidence. Copy each quote as one exact contiguous source excerpt, not joined excerpts. Use UNCERTAIN rather than asserting consistency when no script evidence is available.
\par\medskip
Each CONTRADICTED review must be addressed by an edit; edits may refer only to CONTRADICTED IDs. Do not rewrite abstract descriptions while leaving known invalid commands unchanged. Use at most eight nonoverlapping line-addressed edits. Copy old\_\allowbreak{}text exactly from those lines, without line-number prefixes. Edit only within the Markdown block spans (line through end\_\allowbreak{}line) of the cited invocation IDs; every cited span must intersect the edit. Inline spans include their explanatory paragraph. Do not remove workflows or weaken public requirements. Do not edit protected Required Use Case markers or their contents.
\par\medskip
The program validates references, not behavioral correctness. No task tests, gold, oracle or reward are provided. Keep reasons and summary concise; summary length never determines patch acceptance. Call submit\_\allowbreak{}package\_\allowbreak{}contradiction\_\allowbreak{}assessment once.
\end{framed}

\noindent\begin{minipage}{\linewidth}
\paragraph{Input template.}
\begin{Verbatim}[fontsize=\small,breaklines=true,breakanywhere=true,frame=single]
PUBLIC_REQUEST
<maintenance_request>
INVOCATION_INVENTORY
<numbered_document_invocations>
PYTHON_INTERFACE_FACTS
<extracted_python_interfaces>
SKILL.md (one-based line numbers)
<line_numbered_markdown>
PUBLIC_SCRIPTS
<assembled_script_sources>
\end{Verbatim}
\end{minipage}

\FloatBarrier
\Needspace{12\baselineskip}
\section{Baseline Prompt Templates}
\label{app:baseline-prompts}

The following templates reproduce the system and user instructions for the
three baselines. Each input block follows its user instructions;
angle-bracketed fields denote task-specific content.

\subsection{Markdown-only}
\label{app:prompt-md-baseline}

Both stages use the shared system message below. Each user message consists
of the stage-specific text, followed by the shared editing instruction and
the corresponding input block.

\par\medskip
\Needspace{16\baselineskip}
\begin{framed}
\small\raggedright\interlinepenalty=10000\parskip=0pt
\textbf{Shared instructions.}\par\medskip
\textbf{System message}\par\nobreak\smallskip\nobreak
You audit and repair one visible Agent Skill package. Call submit\_\allowbreak{}skill\_\allowbreak{}patch exactly once, including an empty edits list when no change is justified. Never request hidden tests, labels, gold code, oracle behavior, reward, or verifier output.
\par\medskip
\textbf{Shared editing instruction}\par\nobreak\smallskip\nobreak
Call submit\_\allowbreak{}skill\_\allowbreak{}patch exactly once. Return zero to eight bounded line edits. Every edit must copy path and sha256 from VISIBLE\_\allowbreak{}PACKAGE into path and expected\_\allowbreak{}file\_\allowbreak{}sha256. numbered\_\allowbreak{}content line prefixes are display-only and must not be copied into replacement. replace\_\allowbreak{}lines replaces the inclusive start\_\allowbreak{}line..end\_\allowbreak{}line range; insert\_\allowbreak{}before and insert\_\allowbreak{}after require equal start\_\allowbreak{}line and end\_\allowbreak{}line. Use an empty target\_\allowbreak{}node\_\allowbreak{}id unless a supplied structural packet identifies the exact node. Do not return a unified diff or complete files. Use zero edits when no change is needed.
\end{framed}
\par

\par\medskip
\Needspace{16\baselineskip}
\begin{framed}
\small\raggedright\interlinepenalty=10000\parskip=0pt
\textbf{Initial editing.}\par\medskip
\textbf{Stage-specific user message}\par\nobreak\smallskip\nobreak
Audit and repair one visible executable Agent Skill package. Do not assume a defect exists. Preserve unrelated behavior and public interfaces. Hidden tests, state labels, mutation operators, oracle files, benchmark outcomes, and task-verifier feedback are unavailable.
\par\smallskip
Only Markdown files are visible and editable; executable files are unavailable and immutable.
\par\medskip
\noindent\begin{minipage}{\linewidth}
\textbf{Input template}\par\smallskip
\begin{Verbatim}[fontsize=\footnotesize,breaklines=true,breakanywhere=true]
PUBLIC_TASK.json
<public_task_json>
USER_REQUEST.md
<maintenance_request>
VISIBLE_PACKAGE.json
<visible_markdown_files_json>
\end{Verbatim}
\end{minipage}
\end{framed}
\par

\par\medskip
\Needspace{16\baselineskip}
\begin{framed}
\small\raggedright\interlinepenalty=10000\parskip=0pt
\textbf{Revision.}\par\medskip
\textbf{Stage-specific user message}\par\nobreak\smallskip\nobreak
Revise one frozen executable-skill package proposal. Inspect it again against the public request and the generic checks. Do not assume another edit is necessary. Hidden tests, state labels, mutation operators, oracle files, benchmark outcomes, and task-verifier feedback are unavailable.
\par\smallskip
Only Markdown remains visible and editable; scripts are unavailable and immutable.
\par\medskip
\noindent\begin{minipage}{\linewidth}
\textbf{Input template}\par\smallskip
\begin{Verbatim}[fontsize=\footnotesize,breaklines=true,breakanywhere=true]
USER_REQUEST.md
<maintenance_request>
GENERIC_PUBLIC_CHECKS.json
<syntax_and_edit_scope_checks_json>
FROZEN_PROPOSAL_PACKAGE.json
<revised_markdown_files_json>
\end{Verbatim}
\end{minipage}
\end{framed}
\par

\Needspace{12\baselineskip}
\subsection{Raw Package}
\label{app:prompt-raw-baseline}

\par\medskip
\Needspace{16\baselineskip}
\begin{framed}
\small\raggedright\interlinepenalty=10000\parskip=0pt
\textbf{Package editing.}\par\medskip
\textbf{System message}\par\nobreak\smallskip\nobreak
You are a precise maintainer of executable Agent Skills.
\par\medskip
\textbf{User message}\par\nobreak\smallskip\nobreak
Repair the complete visible executable Agent Skill package so it satisfies the public user request. Inspect SKILL.md and every relevant script before deciding. The package may contain one or several connected defects; repair the complete caller-helper and cross-file closure. Keep documentation and scripts consistent, preserve unrelated behavior and public interfaces, and edit only SKILL.md or existing files below scripts/. If no change is needed, submit an empty edits list.
\par\smallskip
No AST, structural location, hidden tests, labels, verifier output, gold, oracle, reward, or prior benchmark outcome is available.
\par\smallskip
Use submit\_\allowbreak{}skill\_\allowbreak{}patch exactly once. For scripts use replace\_\allowbreak{}lines with inclusive display line ranges; for SKILL.md use replace\_\allowbreak{}lines or append\_\allowbreak{}markdown. Copy each visible file sha256 into expected\_\allowbreak{}file\_\allowbreak{}sha256. Leave node identifiers and node hashes empty. Display-only '\textless{}line\textgreater{}: ' prefixes must not enter replacements. Return bounded edits, not unified diffs or complete files.
\par\medskip
\noindent\begin{minipage}{\linewidth}
\textbf{Input template}\par\smallskip
\begin{Verbatim}[fontsize=\footnotesize,breaklines=true,breakanywhere=true]
PUBLIC_TASK.json
<public_task_json>
PUBLIC_REQUEST.md
<maintenance_request>
VISIBLE_PACKAGE.json
<visible_package_files_json>
\end{Verbatim}
\end{minipage}
\end{framed}
\par

\Needspace{12\baselineskip}
\subsection{CoEvoSkills}
\label{app:prompt-coevo-baseline}

\par\medskip
\Needspace{16\baselineskip}
\begin{framed}
\small\raggedright\interlinepenalty=10000\parskip=0pt
\textbf{Surrogate verification.}\par\medskip
\textbf{System message}\par\nobreak\smallskip\nobreak
You are the isolated Surrogate Verifier in a CoEvoSkills-style loop. Inspect only the immutable public task and package. Call submit\_\allowbreak{}surrogate\_\allowbreak{}report exactly once and never emit or apply a patch.
\par\medskip
\textbf{User message}\par\nobreak\smallskip\nobreak
Independently verify one immutable executable Agent Skill package. The package is the candidate artifact, so inspecting its public files is required; generator reasoning and prior conversations are unavailable. Derive requirements only from PUBLIC\_\allowbreak{}TASK and PUBLIC\_\allowbreak{}REQUEST. Build a small declarative surrogate test suite using only the allowed check kinds. Exact text checks must encode a requirement stated by the request, not a guessed hidden answer. Set FAIL only when public evidence supports a concrete defect; set PASS when the visible package satisfies all checkable requirements; use ABSTAIN when the request cannot be checked from the visible package. Do not propose edits and do not ask for hidden tests, mutation labels, gold code, oracle behavior, benchmark outcomes, or reward. Call submit\_\allowbreak{}surrogate\_\allowbreak{}report exactly once.
\par\medskip
\noindent\begin{minipage}{\linewidth}
\textbf{Input template}\par\smallskip
\begin{Verbatim}[fontsize=\footnotesize,breaklines=true,breakanywhere=true]
PUBLIC_TASK.json
<public_task_json>
PUBLIC_REQUEST.md
<maintenance_request>
IMMUTABLE_VISIBLE_PACKAGE.json
<visible_package_files_json>
\end{Verbatim}
\end{minipage}
\end{framed}
\par

\par\medskip
\Needspace{16\baselineskip}
\begin{framed}
\small\raggedright\interlinepenalty=10000\parskip=0pt
\textbf{Generator revision.}\par\medskip
\textbf{System message}\par\nobreak\smallskip\nobreak
You are the Skill Generator performing exactly one refinement from an independent surrogate report. Call submit\_\allowbreak{}skill\_\allowbreak{}patch exactly once.
\par\medskip
\textbf{User message}\par\nobreak\smallskip\nobreak
Perform the single allowed Skill Generator refinement in a CoEvoSkills-style loop. The independent surrogate report is a fallible public-evidence hypothesis, not an oracle: recheck it against the request and package before editing.
\par\smallskip
Make the smallest coherent repair. Edit only SKILL.md or files under scripts/. Preserve unrelated behavior and public interfaces. You may create a missing text source file under scripts/, but may not delete or rename files.
\par\smallskip
Call submit\_\allowbreak{}skill\_\allowbreak{}patch exactly once with zero to eight bounded edits; use zero edits if the surrogate diagnosis is not justified. For an existing file, use replace\_\allowbreak{}exact, copy its visible sha256, and copy a nonempty, unique, verbatim old snippet into observed\_\allowbreak{}source. The old snippet and replacement must each cover the complete syntactic construct being changed (for example the entire function, branch, mapping entry, or command block), so no orphaned header, else branch, delimiter, or indentation remains. For create\_\allowbreak{}file use an empty expected\_\allowbreak{}file\_\allowbreak{}sha256 and empty observed\_\allowbreak{}source.
\par\smallskip
Hidden tests, labels, gold code, oracle behavior, benchmark outcomes, and task-verifier feedback remain unavailable.
\par\medskip
\noindent\begin{minipage}{\linewidth}
\textbf{Input template}\par\smallskip
\begin{Verbatim}[fontsize=\footnotesize,breaklines=true,breakanywhere=true]
PUBLIC_TASK.json
<public_task_json>
PUBLIC_REQUEST.md
<maintenance_request>
SURROGATE_REPORT.json
<surrogate_report_json>
EXECUTED_SURROGATE_CHECKS.json
<executed_surrogate_checks_json>
CURRENT_VISIBLE_PACKAGE.json
<visible_package_files_json>
\end{Verbatim}
\end{minipage}
\end{framed}
\par

\FloatBarrier
\Needspace{10\baselineskip}
\section{Limitations}
\label{sec:limitations}

The In-the-Wild track is predominantly Python, and both tracks use constructed
faults to evaluate package maintenance. Broader language coverage and tasks
derived from naturally occurring failures would help assess generalization
to the wider Skill ecosystem. Our evaluation measures package-level correctness
and documentation--script consistency; reliability in long-horizon downstream
Agent workflows remains to be evaluated.

\end{document}